\documentclass[accepted]{uai2023} 

\usepackage[american]{babel}

\usepackage{natbib} 
\usepackage{mathtools} 
\usepackage{booktabs} 
\usepackage{tikz} 
\usepackage{amsmath}
\usepackage{amssymb}
\usepackage{graphicx}
\usepackage{hyperref}
\usepackage{mathtools}
\usepackage{mathrsfs}
\usepackage{xcolor}

\title{Amortized Inference for Gaussian Process Hyperparameters of Structured Kernels}

\author{Matthias Bitzer}
\author{Mona Meister}
\author{Christoph Zimmer}

\affil{%
Bosch Center for Artificial Intelligence\\ Renningen, Germany
}

\begin{document}
\maketitle

\begin{abstract}
Learning the kernel parameters for Gaussian processes is often the computational bottleneck in applications such as online learning, Bayesian optimization, or active learning. Amortizing parameter inference over different datasets is a promising approach to dramatically speed up training time. However, existing methods restrict the amortized inference procedure to a fixed kernel structure. The amortization network must be redesigned manually and trained again in case a different kernel is employed, which leads to a large overhead in design time and training time. We propose amortizing kernel parameter inference over a complete \textit{kernel-structure-family} rather than a fixed kernel structure. We do that via defining an amortization network over pairs of datasets and kernel structures. This enables fast kernel inference for each element in the kernel family without retraining the amortization network. As a by-product, our amortization network is able to do fast ensembling over kernel structures. In our experiments, we show drastically reduced inference time combined with competitive test performance for a large set of kernels and datasets.
\end{abstract}
\section{Introduction}
Gaussian processes (GPs) are an important class of models that can be used in a wide range of tasks such as Bayesian optimization \citep{PracticalBO,ChemBO}, active learning \citep{algpsshapecontrol,SafeALTimeSeries,SafeALMOGP,pmlr-v206-bitzer23a}, or regression \citep{CKS}. Introducing an inductive bias for GPs is achieved by specifying the kernel structure. For example, smoothness, nonstationarity or periodicity can be induced very elegantly by configuring the corresponding kernel. 

Learning the kernel parameters is often a major computational bottleneck and is usually done via marginal likelihood maximization, also called \textit{Type-2-ML}, or via evidence-lower-bound maximization (ELBO) in sparse GP's \citep{svgp}. These methods often require hundreds of optimization steps to learn the kernel parameters. In \cite{AHGP}, this problem is circumvented by using amortized inference \citep{amortized_infer1,amortized_infer2} to predict the kernel parameters via a neural network in one step. This leads to a dramatic reduction in inference time for medium-sized datasets.

However, the method of \cite{AHGP} defines the amortization only for a fixed kernel structure. Importantly, specifying the kernel structure is a crucial design choice for GP's and is often used to induce prior knowledge of the task at hand such as smoothness, nonstationarity, linearity or periodicity. 
In case a different kernel should be used the network in \cite{AHGP} would need to be redesigned and retrained, which is a time-consuming and costly task, considering the vast space of possible kernel structures. We therefore propose amortizing the kernel inference for GP's over the combined space of kernel structures and datasets.

 We define an amortization neural network that gets as input a complete dataset and a symbolical description of the kernel, based on the kernel grammar \citep{CKS}, and outputs the learned kernel parameters. We design the neural network explicitly to cope with the natural invariances of the underlying spaces. Here, we make use of the transformer architecture \citep{AttentionIsAllYouNeed} and its equivariance properties \citep{SetTransformer}. We empirically show that our method leads to a drastic decrease in inference time, while delivering competitive predictive results on real-world datasets. Additionally, we illustrate the generality of our method via defining a fast ensembling over kernel structures that explicitly leverages our architecture. In short, our contributions are
 \begin{enumerate}
 	\item We construct an amortization neural network that is defined on the combined space of kernel structures and datasets. We explicitly incorporate invariances and equivariances of the underlying spaces in the architecture.
 	\item We empirically demonstrate the effectiveness of the amortization over several simulated and real world datasets and kernel structures.
 	\item We show the generality of our approach by enabling a fast ensembling over kernel structures.
 \end{enumerate}
We provide accompanying code at \url{https://github.com/boschresearch/Amor-Struct-GP}.

\section{Background}
In the following section we give necessary background information about Gaussian processes with a focus on the hyperparameter optimization involved. We start by introducing the standard technique to hyperparameter inference for GP's and will then consider amortized inference over multiple datasets, as proposed in \cite{AHGP}. Finally, we consider a broad kernel space over which we will define our proposed \textit{combined} amortization scheme.
\paragraph{Gaussian processes.}
Let $\mathcal{X}\subset \mathbb{R}^{d}$ be the input space for some $d\in\mathbb{N}$.
A Gaussian Process defines a distribution over mappings $f:\mathcal{X}\to \mathbb{R}$ and is fully specified via a positive-definite kernel function $k: \mathcal{X}\times \mathcal{X}\to \mathbb{R}$ and a mean function $m:\mathcal{X}\to \mathbb{R}$. It is characterized by the property, that, for any finite selection of input points $\mathbf{X}=\{x_{1},\dots,x_{n}\} \subset \mathcal{X}$ and any $n\in\mathbb{N}$, the collection of function evaluations $(f(x_{1}),\dots,f(x_{n}))^{\intercal}$ is multivariate Gaussian with mean $m(\mathbf{X}):=(m(x_{1}),\dots,m(x_{n}))^{\intercal}$ and covariance matrix $k(\mathbf{X},\mathbf{X}):=[k(x_{i},x_{j})]_{i,j=1,\dots,n}$. We write $f\sim \mathcal{GP}(m,k)$ to denote that the function $f$ is drawn from a Gaussian process.

Let $\mathcal{D}=\{(x_{i},y_{i})\in\mathbb{R}^{d+1}, i=1,\dots,n\}$ be a dataset for which we want to do regression. The typical modeling assumption for Gaussian process regression presumes a latent function $f\sim \mathcal{GP}(m,k_{\theta})$ with a Gaussian likelihood, thus, $y_{i}=f(x_{i})+\epsilon_{i}$ with $\epsilon_{i} \in \mathcal{N}(0,\sigma^{2})$. The kernel is parameterized with $\theta\in \Theta \subset \mathbb{R}^{p}$ and the complete parameter vector of the GP, including the likelihood variance $\sigma^{2}$, is given with $\phi=(\theta,\sigma^{2}) \in \Phi \subset \mathbb{R}^{p+1}$. An important property of this model is that the marginal likelihood, marginalized over the latent function $f$, can be computed analytically with
\begin{align}
\label{marg_lik}
p(\mathbf{y}|\mathbf{X},\theta,\sigma^{2})&=\int p(\mathbf{y}|f,\mathbf{X},\sigma^{2})p(f|\mathbf{X},\theta)df \\&= \mathcal{N}(\mathbf{y};m(\mathbf{X}),k_{\theta}(\mathbf{X},\mathbf{X})+\sigma^{2}\mathbf{I}). \nonumber
\end{align}
Inference of the inner parameters, which in this case is the infinite dimensional function $f$, can be done analytically. For the outer hyperparameters $\theta$ and $\sigma^2$ the classical way of training is maximizing the marginal-likelihood (\ref{marg_lik}), also called \textit{type-2 maximum likelihood}. Thus, we want to solve the following optimization problem
\begin{align}
\label{optproblem}
(\theta_{*},\sigma^{2}_{*})=\arg\max_{(\theta,\sigma^{2})\in\Phi }\mathrm{log}~p(\mathbf{y}|\mathbf{X},\theta,\sigma^{2})
\end{align}
for a given dataset $\mathcal{D}$. The optimization problem is usually solved via gradient-based optimizers like Adam or L-BFGS. Each step in the optimizer requires a calculation of the marginal-likelihood, which scales cubically in $n$. Furthermore, several hundred optimization steps might be necessary to reach convergence and, depending on the kernel and dataset, multiple restarts are necessary as the optimization problem is non-convex and might end up in a local maxima. In the next section, we will consider an alternative approach of solving (\ref{optproblem}) that only requires one forward-pass through an amortization network.

\paragraph{Parameter amortization.}
\citet{AHGP} presented an alternative method for learning the GP hyperparameters based on amortizing the inference over multiple datasets. In this method, parameter inference for GP's reduces to a prediction via an amortization neural network. The amortization network $g_{\psi}: \mathscr{D} \mapsto \Theta$ with weights $\psi$ is defined on the set of all datasets $\mathscr{D}$ meaning that for any $n\in \mathbb{N}$ and $d \in \mathbb{N}$ the dataset $\mathcal{D}=\{(x_{i},y_{i})\in\mathbb{R}^{d+1},i=1,\dots,n\}$ is part of the input set of the network, thus $\mathcal{D}\in \mathscr{D}$. The output-space is the parameter space $ \Theta$ of the respective kernel for which amortized inference should be done. In case of \citet{AHGP}, the Spectral Mixture Product (SMP) kernel is used, which consists of parameters $\theta_{j}=\{\{w_{m,j}\}_{m=1}^{M},\{\mu_{m,j}\}_{m=1}^{M},\{\sigma^{2}_{m,j}\}_{m=1}^{M}\}$ in the $j$-th dimension for some fixed $M$. The amortization network is designed via consecutive transformer blocks such that it can handle different input sizes and input dimensions of the respective dataset. The network is trained on a dataset of (synthetic) datasets $\{\mathcal{D}^{(l)}\}_{l=1}^{L}\subset \mathscr{D}$ via minimization of the mean average negative marginal likelihood of the datasets. After training, the network is used for one-shot prediction of the kernel parameters $\hat{\theta}=g_{\psi}(\mathcal{D}^{*})$ on an unseen dataset $\mathcal{D}^{*}$. In \cite{AHGP}, the kernel structure is fixed to the one of the SMP kernel. To use it with a different kernel, the network needs to be redesigned and retrained. For example, \citet{rehn2022amortized} changed the architecture to cope with the RBF kernel. Our goal is to do amortized inference over the \textit{combined space of datasets and kernel structures}, which drastically reduces redesign and retrain time and enables fast inference for many existing kernel structures via only one neural network.

Our amortization network consists of a dataset encoder, that is inspired by the architecture of \citet{AHGP} and a novel kernel encoder-decoder block that enables amortization over the combined input of kernel structure and dataset. Both blocks are designed to capture the natural invariances of the underlying structure.
\paragraph{Kernel space.}
\label{kernel_space}
Our goal is to define an amortization procedure over a family of kernels. To be more precise, we consider a family of \textit{structural forms of kernels}. The basis for this kernel space is the kernel grammar presented in \citet{CKS}. Here, each kernel structure is expressed as a symbolic expression $\mathcal{S}$ made of base symbols $\mathcal{B}$. The base symbols might include simple elementary kernels like the Squared-Exponential kernel represented as the symbol $\textrm{SE}$, the linear kernel as $\textrm{LIN}$ or the periodic kernel as \textrm{PER}. More complex expressions can be formed with multiplication and addition of base kernels/symbols. For example, one might construct a more complex structural form of a kernel via the expression $\mathrm{SE}\times \mathrm{LIN} + \mathrm{PER}$. Each expression $\mathcal{S}$ describes a structural form of a kernel - thus, the mathematical equation that governs the associated kernel $k$. The base kernels and therefore the combined expressions come with parameters $\theta$ and thus each expression has its own associated parameter space $\Theta_{\mathcal{S}}$.

The kernel grammar in \cite{CKS} considers all possible algebraic expressions of the base kernels. We consider a subset of the kernel grammar that leads to a rich kernel space on the one hand and one that can be easily represented in a neural network on the other hand.

First, we define a set of base symbols $\mathcal{B}$ which consists of a set of elementary kernels like $\mathrm{SE}, \mathrm{LIN}$ and $\mathrm{PER}$ and its two-gram multiplications like $\mathrm{SE}\times\mathrm{LIN}$, $\mathrm{SE}\times\mathrm{PER}$ and $\mathrm{LIN}\times\mathrm{PER}$. This is a similar symbol set as used in \cite{KernelIdentWithTrafo}. All base symbols are defined on single dimensions, and we denote the concrete dimension via an index, e.g. $\mathrm{SE}_{i}$ for the Squared-Exponential kernel on dimension $i$ and summarize the sets of indexed base symbols to $\mathcal{B}^{(i)}$. Our kernel space is then defined as an addition of base symbols within the dimension and a multiplication over dimensions:
$
\mathcal{S}=\prod_{i=1}^{d}\sum_{j=1}^{N_{i}} \mathcal{S}_{i,j}$ with $S_{i,j}\in \mathcal{B}^{(i)}.
$
For example, the following kernel would be part of the complete kernel space:
\begin{align}
\label{example_expression}
\overbrace{
	\underbrace{(\underbrace{\mathrm{SE}_{1}\times \mathrm{LIN}_{1}}_{\textcolor{violet}{\text{symbol of }\mathcal{B}}} +~ \mathrm{SE}_{1})}_{\textcolor{blue}{\text{Addition within dimension}}} \times ~(\mathrm{SE}_{2} + \mathrm{PER}_{2})}^{\textcolor{red}{\text{Multiplication over dimensions}}}.
\end{align} 
We denote the complete kernel space with $\mathcal{K}$. This kernel space contains popular kernels like the $\textrm{ARD-RBF}$ kernel with $\prod_{i=1}^{d}\textrm{SE}_{i}$ or the $d$-dimensional periodic kernel with $\prod_{i=1}^{d}\mathrm{PER}_{i}$. Additionally, kernels that act differently on different dimensions are included in the kernel space. 

Depending on the kernel expression, the parameter space $\Theta_{\mathcal{S}}$ can vary significantly in dimensionality. For example, the \textrm{ARD-RBF} kernel $\mathcal{S}=\prod_{i=1}^{d}\textrm{SE}_{i}$ on dimension $d$ contains one lengthscale and variance\footnote{We use the parameterization of the base kernels from the kernel grammar \cite{CKS}. Here each base kernel in each dimension has its own variance.} parameter per dimension, such that $\Theta_{\mathcal{S}} \subset \mathbb{R}^{2d}$. The d-dimensional periodic kernel $\mathcal{S}=\prod_{i=1}^{d}\mathrm{PER}_{i}$ contains an additional feature specific period parameter such that $\Theta_{\mathcal{S}} \subset \mathbb{R}^{3d}$. Thus, being able to deal with different sizes of parameter spaces will be important for our proposed amortization scheme.
\section{Method}
We propose amortizing the kernel inference for GP's over the combined space of datasets and kernel structures. This enables fast inference for many kernels, as well as fast ensembling. For this, we construct an amortization network $(\mathscr{D},\mathcal{K}) \ni (\mathcal{D},\mathcal{S}) \mapsto g_{\psi}(\mathcal{D},\mathcal{S}) \in \Theta_{\mathcal{S}}\cup \mathbb{R}_{+}$ that maps from the combined space of datasets $\mathscr{D}$ and kernel structures $\mathcal{K}$ to the parameter space of the GP with respective kernel $\Theta_{\mathcal{S}}\cup \mathbb{R}_{+}$. The trained network is then used to one-shot predict GP parameters $(\hat{\theta}_{\mathcal{S}},\hat{\sigma}^{2})=g_{\psi}(\mathcal{D}^{*},\mathcal{S})$ of the specified kernel structure $\mathcal{S}$ for an unseen dataset $\mathcal{D}^{*}$. We denote with $\psi$ the (trainable) parameters of the amortization network. In the following subsections, we describe the architecture of the network and the learning procedure of $g_{\psi}$.
\subsection{Architecture}

\begin{figure*}[t]
	\centering
	\includegraphics[width=0.99\linewidth]{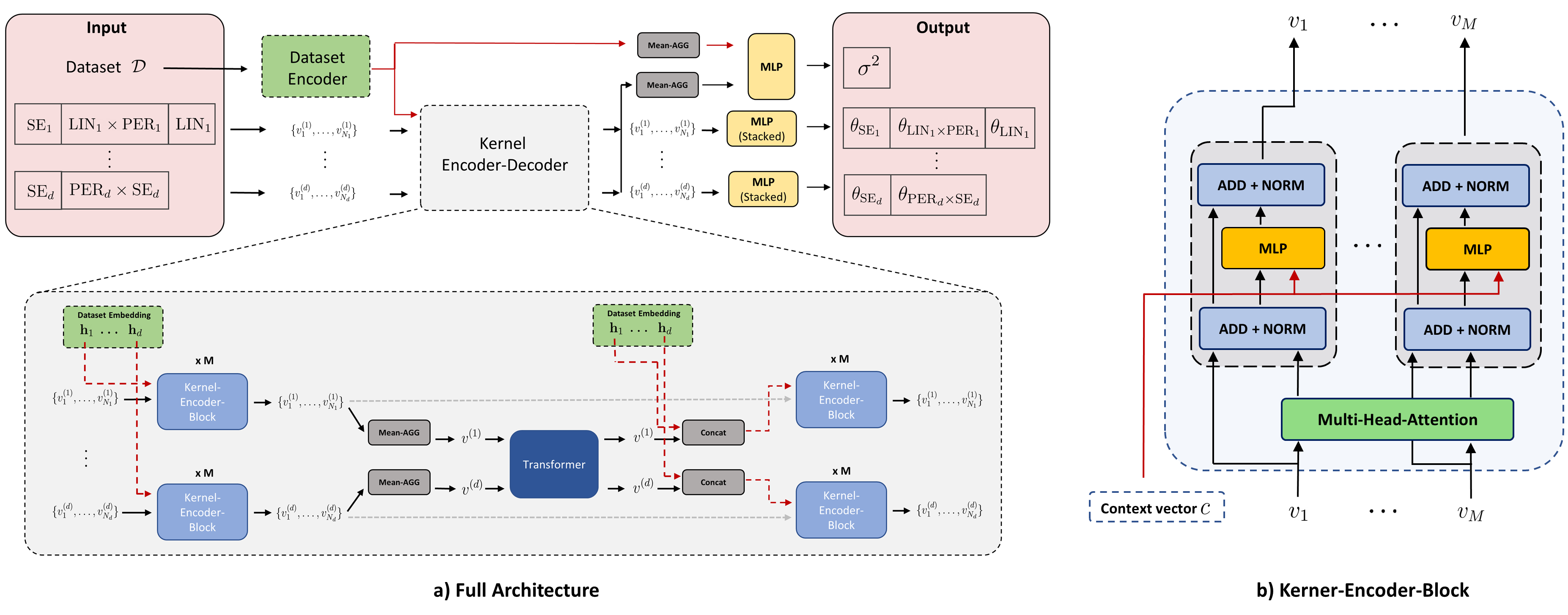}
	\caption{a) Illustration of the full amortization network, which gets as input the dataset $\mathcal{D}$ and a kernel expression $\mathcal{S}$ decomposed in a sequence of sequences of base symbols and outputs the kernel hyperparameters $\theta$ in the respective parameter space $\Theta_{S}$. b)  The main layer used in the Kernel-Encoder-Decoder. It gets as input a sequence of vectors and a context vector and outputs a transformed sequence of vectors. The context vector enters the MLP layer.}
	\label{fig:architectureoverview}
\end{figure*}

The model gets as input a dataset $\mathcal{D}=\{(x_{i},y_{i}) \in \mathbb{R}^{d+1}| i=1,\dots,n\}$ where $n \in \mathbb{N}$ is the number of datapoints and $d \in \mathbb{N}$ is the number of input dimensions. Additionally, the model receives the kernel expression $\mathcal{S}$ as input. As $\mathcal{S}$ is a multiplication over dimension-wise sub-expressions $\mathcal{S}_{i}$, we can represent the expression $\mathcal{S}$ as a sequence of its sub-expressions $[\mathcal{S}_{1},\dots,\mathcal{S}_{d}]$. Similarly, we can decompose the expressions in each dimension into a sequence of base symbols. Thus, we represent/store the expression as a sequence of sequences of base symbols $\bigg[[B_{1}^{(i)},\dots,B^{(i)}_{N_{i}}]|i=1,\dots,d;B_{j}^{(i)}\in \mathcal{B}\bigg]$. We encode each base symbol via one-hot-encoding such that the sub-expression in each dimension $\mathcal{S}_{i}$ is represented via a sequence of vectors $\mathcal{V}_{i}=[v_{1}^{(i)},\dots,v_{N_{i}}^{(i)}]$ with $v_{j}^{(i)}\in \mathbb{R}^{|\mathcal{B}|}$. The whole expression is then represented via $\mathcal{V}_{\mathcal{S}}=[\mathcal{V}_{1},\dots,\mathcal{V}_{d}]$. 

Our architecture consists of three main parts, the dataset encoder $g_{D}$, that takes as input the dataset $\mathcal{D}$ and returns a sequence of dimension-wise embeddings $\mathbf{h}_{\mathcal{D}}=[\mathbf{h}_{1},\dots,\mathbf{h}_{d}]$, a kernel encoder-decoder $g_{k}(\mathbf{h}_{\mathcal{D}},\mathcal{V}_{\mathcal{S}})$ that gets as input a sequence of sequences of encoded base symbols $[\mathcal{V}_{1},\dots,\mathcal{V}_{d}]$ and the dataset embeddings $[\mathbf{h}_{1},\dots,\mathbf{h}_{d}]$ and outputs a transformed sequence of sequences of kernel embeddings $[\mathcal{V}_{1},\dots,\mathcal{V}_{d}]$ and finally an output layer that maps the kernel embeddings to the respective parameter space of the base kernels.

 We design the different parts of the architecture to cope with several symmetries. Similar to \cite{AHGP}, our network is permutation invariant to the shuffling of the dataset elements. Furthermore, the dataset encoder and the kernel encoder-decoder are equivariant to the permutation of input dimensions. Incorporating these symmetries enables generalization to datasets with sizes and input dimensions that were not present in the training phase. Lastly, the final output is invariant to a shuffling of the base symbols in each dimension, which is important as the sequences describe additions. The prediction of the network is thus not dependent on the order in which the additions are represented in $\mathcal{S}$.
 
We present the single parts of the architecture in the following paragraphs and an overview in Figure \ref{fig:architectureoverview} a).
\paragraph{Dataset-Encoder.}
The dataset encoder takes as input the dataset $\mathcal{D}$ and returns a sequence of dimension embeddings  $[\mathbf{h}_{1},\dots,\mathbf{h}_{d}]$. We utilize the encoder part of the Transformer architecture without positional encoding \citep{AttentionIsAllYouNeed,SetTransformer} in multiple parts of our architecture and refer to it as a $\textbf{Transformer}$ block. 
Each block maps a sequence of vectors to a sequence of transformed vectors $[a_{1},\dots,a_{1}]\leftarrow \textbf{Transformer}([a_{1},\dots,a_{1}])$ using multiple multi-head-self-attention layers \citep{AttentionIsAllYouNeed}. We consecutively apply $\textbf{Transformer}$ blocks to different hidden embeddings to construct a sequence of dimension embeddings  $[\mathbf{h}_{1},\dots,\mathbf{h}_{d}]$ with $\mathbf{h}_{i}\in\mathbb{R}^{2h}$. It involves the following steps:
\begin{enumerate}
	\item The dataset is divided into dimension-wise sequences $[(x_{j}^{(i)},y_{j})]_{j=1}^{n}$ where $x_{j}^{(i)}$ is the $i$-th dimension of point $x_{j}$.
	\item Each sequence $[(x_{j}^{(i)},y_{j})]_{j=1}^{n}$ is mapped element-wise via a linear layer to construct a sequence of embeddings per dimension $[h_{1}^{(i)},\dots,h_{n}^{(i)}]$ with $h_{j}^{(i)}\in\mathbb{R}^{h}$.
	\item Each sequence $[h_{j}^{(i)}]_{j=1}^{n}, i=1,\dots,d$ is given to a $\textbf{Transformer}$ block (shared over the $d$ sequences) that outputs a transformed sequence $[h_{j}^{(i)}]_{j=1}^{n}$.
	\item So far, each datapoint was only able to attend to other datapoints inside its dimension. In a next step, we create per datapoint embeddings via mean aggregation $h_{j}=\textbf{MeanAGG}([h_{j}^{(1)},\dots,h_{j}^{(d)}])$ leading to a sequence of datapoint embeddings $[h_{j}]_{j=1}^{n}$ with $h_{j}\in\mathbb{R}^{h}$.
	\item The datapoint embeddings are put into a \textbf{Transformer} block to form a transformed sequence of datapoint embeddings $[h_{1},\dots,h_{n}]$.
	\item In order to construct embeddings per dimension again, we append the datapoint embedding to the sequences of embeddings of step 3, thus $h_{j}^{(i)}\leftarrow\mathbf{Concat}(h_{j},h_{j}^{(i)})$ which results in sequences $[h_{1}^{(i)},\dots,h_{n}^{(i)}]$ with $h_{j}^{(i)}\in\mathbb{R}^{2h}$. 
	\item Each updated sequence $[h_{1}^{(i)},\dots,h_{n}^{(i)}], i=1,\dots,d$ is again given to a (shared) $\textbf{Transformer}$ block that outputs a transformed sequence $[h_{1}^{(i)},\dots,h_{n}^{(i)}]$.
	\item  In order to get dimension embeddings, we aggregate the sequence via mean aggregation to $\mathbf{h}_{i}=\textbf{MeanAGG}([h_{1}^{(i)},\dots,h_{n}^{(i)}])$ leading to a sequence of dimension embeddings $[\mathbf{h}_{i}]_{i=1}^{d}$ with $\mathbf{h}_{i}\in \mathbb{R}^{2h}$.
	\item The sequence of dimension embeddings $[\mathbf{h}_{i}]_{i=1}^{d}$ is again put through a $\textbf{Transformer}$ block to get a sequence of dimension embeddings $[\mathbf{h}_{i}]_{i=1}^{d}$ that contains shared information across dimensions.
\end{enumerate}
The encoder is very similar to the one in \citet{AHGP}. The only difference are the steps 3.~to 6. We incorporate these steps to prevent the permutation invariance to shuffling in the seperated sequences $[(x_{j}^{(i)},y_{j})]_{j=1}^{n}$, which can lead to pathologies (see Appendix C). Our encoder is still invariant to a shuffling of the dataset elements and permutation equivariant to a shuffling of the input dimensions. We give rigorous proofs in Appendix C.

\paragraph{Kernel-Encoder-Decoder.} The kernel encoder-decoder block is meant to translate the structure of the kernel given through $\mathcal{V}_{\mathcal{S}}=[\mathcal{V}_{1},\dots,\mathcal{V}_{d}]$ into transformed embeddings $\mathcal{V}_{\mathcal{S}}=[\mathcal{V}_{1},\dots,\mathcal{V}_{d}]$ that incorporate the information of the dataset and the global information about the kernel structure. These embeddings can be used to predict kernel parameters of the base symbols that are associated with each embedding element $v_{j}^{(i)}$. We call the block encoder-decoder, as the global structure of the expression $\mathcal{S}$ needs to be encoded and then, using the information of the global structure and the dataset, each embedding of base symbols $v_{j}^{(i)}$ needs to be decoded into a vector that contains information about the kernel parameters of the base symbol/kernel.

The main building block is the \textbf{Kernel-Encoder-Block} as shown in Figure \ref{fig:architectureoverview} b). This block maps a context vector $c\in\mathbb{R}^{l}$ and a sequence of vectors $[v_{1},\dots,v_{M}]$ with $v_{j}\in\mathbb{R}^{h}$ to a transformed sequence of vectors $[v_{1},\dots,v_{M}]$ with $v_{j}\in\mathbb{R}^{h}$. It first applies self-attention to the sequence, followed by a concatenation of the context vector to the input of the element-wise multi-layer-perceptron (MLP) layer. Given a context vector $c$, this layer is permutation-equivariant to a shuffling of the sequence.

The \textbf{Kernel Encoder-Decoder} consists of the following steps:
\begin{enumerate}
	\item Each sequence of base kernel embeddings per dimension $[v_{1}^{(i)},\dots,v_{N_{i}}^{(i)}]$ is given to a (shared) stack of \textbf{Kernel-Encoder-Block} layers with the dataset embedding of the respective dimension $\mathbf{h}_{i}$ as context vector. The output is a transformed sequence $[v_{1}^{(i)},\dots,v_{N_{i}}^{(i)}]$. 
	\item So far, only the information of the base kernels along one dimension is shared. Thus, we form dimension-wise kernel embeddings via mean aggregation to $\mathbf{v}_{i}=\textbf{MeanAGG}([v_{1}^{(i)},\dots,v_{n}^{(i)}])$ leading to the sequence $[\mathbf{v}_{1},\dots,\mathbf{v}_{d}]$ with $\mathbf{v}_{i}\in\mathbb{R}^{h}$. Each element in the sequence is an embedding of the kernel inside dimension $i$.
	\item To form a shared (global) representation of the kernel, we put the sequence of dimension-wise kernel embeddings to a \textbf{Transformer} block and receive a transformed sequence of dimension-wise kernel embeddings $[\mathbf{v}_{1},\dots,\mathbf{v}_{d}]$.
	\item Finally, we apply again a (shared) stack of \textbf{Kernel-Encoder-Block} layers to the base kernel embeddings per dimension $[v_{1}^{(i)},\dots,v_{N_{i}}^{(i)}]$ with extended context vector $c_{i}=\mathbf{Concat}(\mathbf{h}_{i},\mathbf{v}_{i})$ such that the shared kernel representation as well as the dataset encoding are part of the context. This gives us the final sequence of kernel embeddings per dimension  $[v_{1}^{(i)},\dots,v_{N_{i}}^{(i)}]$.
\end{enumerate}
\begin{figure*}[t]
	\centering
	\includegraphics[width=0.99\linewidth]{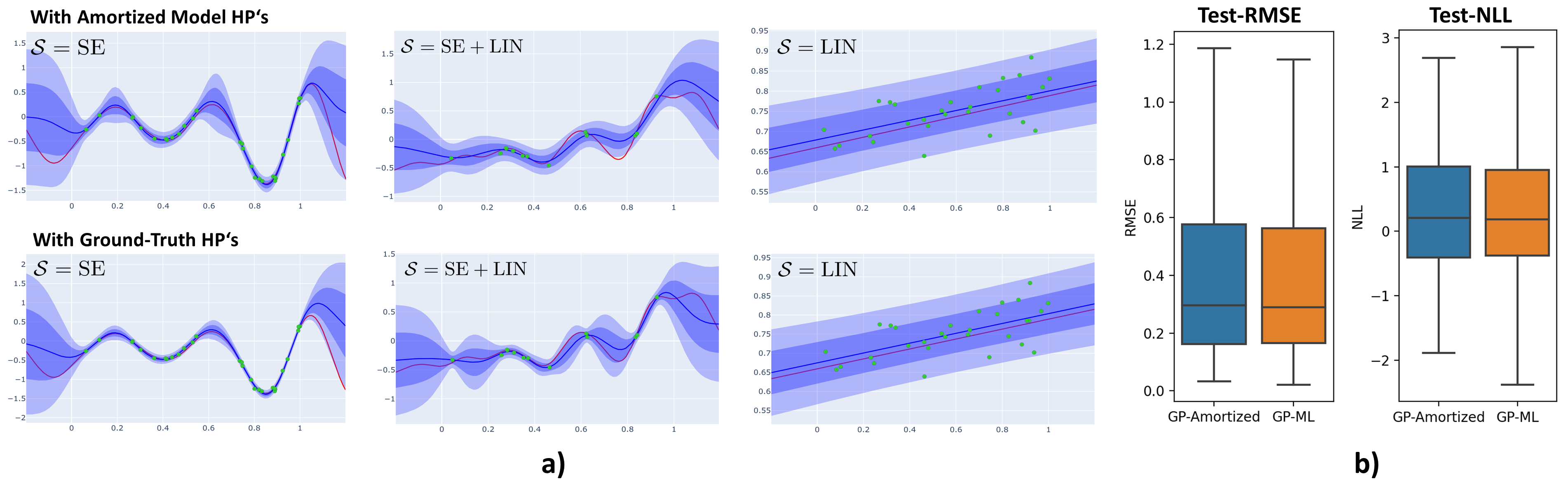}
	\caption{In a) each column contains a ground truth function (red-line) drawn from a GP with kernel $\mathcal{S}$ and hyperparameter $\phi^{*}_{\mathcal{S}}$. Noisy datapoints from the ground truth function are shown in green.  The upper row shows the resulting predictive distribution with predicted GP parameters $\hat{\phi}_{\mathcal{S}}=g_{\psi}(\mathcal{D},\mathcal{S})$ and the lower row shows the predictive distribution with ground truth hyperparameter $\phi^{*}_{\mathcal{S}}$. In b) we show boxplots of the RMSE and NLL scores measured on 200 unseen, simulated dataset-kernel pairs $(\tilde{\mathcal{D}}_{l},\mathcal{S}_{l})$ for our method and for a GP with Type-2-ML inference. The datasets are sampled from the same distribution as used for training.}
	\label{fig:toydata}
\end{figure*}
\paragraph{Output-Layer.}
The final part of the architecture is the prediction head for the kernel parameters. For a given expression $\mathcal{S}$, this layer has as output space the corresponding parameters space $\Theta_{\mathcal{S}}$. It gets as input the kernel embeddings per dimension  $[v_{1}^{(i)},\dots,v_{N_{i}}^{(i)}]$ of the \textbf{Kernel Encoder-Decoder}. Each embedding $v_{j}^{(i)}$ is associated with one base-symbol $B_{j}^{(i)}$ and each base-symbol has its own, fixed parameter space $\Theta_{B_{j}^{(i)}}$, like for example $\Theta_{SE}\subset\mathbb{R}^{2}$. We therefore realize the final mapping to $\Theta_{\mathcal{S}}$ via mapping each symbol related embedding $v_{j}^{(i)}\in\mathbb{R}^{h}$ to the respective parameter space of the base-symbol. We do this via separate \textbf{MLP} blocks for each base-symbol (more details in Appendix C).

In order to get an end-to-end amortization network, we also need a prediction for the likelihood variance. The variance depends on the kernel choice and the dataset. Thus, we form global embeddings of the kernel, via mean aggregation of all kernel embeddings per dimension  $[v_{1}^{(i)},\dots,v_{N_{i}}^{(i)}]$, and of the dataset via mean aggregation of the dimension embeddings  $[\mathbf{h}_{1},\dots,\mathbf{h}_{d}]$. We concatenate both global embeddings and use an \textbf{MLP} block to predict the noise variance.

In summary, our network predicts the kernel parameters and noise variance for a given dataset $\mathcal{D}$ and expression $\mathcal{S}$:
\begin{align}
(\hat{\theta}_{\mathcal{S}},\hat{\sigma}^{2})=g_{\psi}(\mathcal{D},\mathcal{S}).
\end{align}
It accounts for the natural invariances/equivariances of the respective spaces, which we elaborate in Appendix C.

\paragraph{Computational complexity of one forward pass.} One prediction of the kernel parameters via one forward pass scales with $\mathcal{O}(n^{2}+d^{2}+l^{2})$ where $n$ is the number of datapoints and $d$ the number of dimensions in the input dataset $\mathcal{D}$ and $l=\max(N_{1},\dots,N_{d})$ is the maximum number of symbols of the kernel sub-expressions $\mathcal{S}_{i}$ in the dimensions $i=1,\dots,d$. This follows directly from the quadratic complexity (in the sequence length) of the multi-head-self-attention layer. This complexity could be reduced via the usage of sparse attention layers \citep{sparseTransformer}.
\subsection{Training Procedure}
Our objective is to train a general purpose prediction network $g(\mathcal{D},\mathcal{S})$ that can act on, in principle, all (medium-sized) datasets and all expressions $\mathcal{S}$ in the described kernel space. To conquer this challenge with enough data, we train our network purely on simulated datasets. We reflect the variety of inputs via sampling pairs $(\mathcal{D}_{l},\mathcal{S}_{l})$ from a broad distributions $(\mathcal{D}_{l},\mathcal{S}_{l}) \sim p(\mathcal{D},\mathcal{S})$. Given a dataset of sampled dataset-kernel pairs $\{(\mathcal{D}_{l},\mathcal{S}_{l})\}_{l=1}^{L}$, we utilize the average mean negative marginal-likelihood
\begin{align}
\label{main_loss} 
&\mathcal{L}(\psi,\{(\mathcal{D}_{l},\mathcal{S}_{l})\}_{l=1}^{L})\\&=-\frac{1}{L}\sum_{l=1}^{L} \frac{1}{|\mathcal{D}_{l}|} \mathrm{log}~p\bigg(\mathbf{y}_{l}\bigg|\mathbf{X}_{l},(\theta_{l},\sigma_{l}^{2})=g_{\psi}(\mathcal{D}_{l},\mathcal{S}_{l})\bigg)\nonumber
\end{align}
as loss-function. This reflects our goal to train a network that resembles the marginal-likelihood optimization of the kernel hyperparameters for a given kernel structure and a given dataset.

\paragraph{Sampling distribution.} We sample $(\mathcal{D},\mathcal{S})$ using the following scheme (we give a sketch here, details on the utilized distributions/priors can be found in Appendix A). First, we draw the number of input-dimensions $d$ and datapoints $n$. Given $d$ we draw a kernel expression $\mathcal{S}$, where we draw the subexpressions $\mathcal{S}_{i}$ independently of each other. Each subexpression can have a different number of base symbols. Each base-symbol/base-kernel comes with a prior on its hyperparameters. In order to generate a dataset $\mathcal{D}$ that stems from the induced prior in function space of $\mathcal{S}$, we sample from the hyperparameter prior $\theta\sim p_{\mathcal{S}}(\theta)$ with $\theta \in \Theta_{\mathcal{S}}$. We use broad Gamma priors for the kernel parameters. Next, we draw the input set $\mathbf{X}=\{x_{1},\dots,x_{n}\}$ uniformly from $[0,1]^{d}$. Finally, we draw the observations from the GP via $\mathbf{y} \sim \mathcal{N}(\mathbf{0},k_{\mathcal{S},\theta}(\mathbf{X},\mathbf{X})+\sigma^{2}\mathbf{I})$, where $\sigma^{2}\sim p(\sigma^{2})$.

 When constructing the pair $(\mathcal{D},\mathcal{S})$, we distinguish two modes. The first mode is that $\mathcal{D}$ is sampled from the induced prior of  $\mathcal{S}$ - we refer to this mode as the \textit{positive} sample. For the second mode, we sample $\mathcal{D}$ using a different expression $\tilde{\mathcal{S}}$ - we refer to this mode as the \textit{negative} sample. The reason for these two modes is that we cannot assume that only datasets from the induced prior of $\mathcal{S}$ will be used as input to the prediction network. There will always be a misspecification of the kernel. 

\paragraph{Training parameters.} During training, we sample each batch $\{(\mathcal{D}_{l},\mathcal{S}_{l})\}_{l=1}^{L}$ of size $L$ on-the-fly from the sampling distribution $p(\mathcal{D},\mathcal{S})$. This enables processing a huge corpus of dataset-kernel pairs. We employ $\textrm{RAdam}$ \citep{RADAM} as optimizer with a constant lengthscale.
\begin{figure*}[t]
	\centering
	\includegraphics[width=0.99\linewidth]{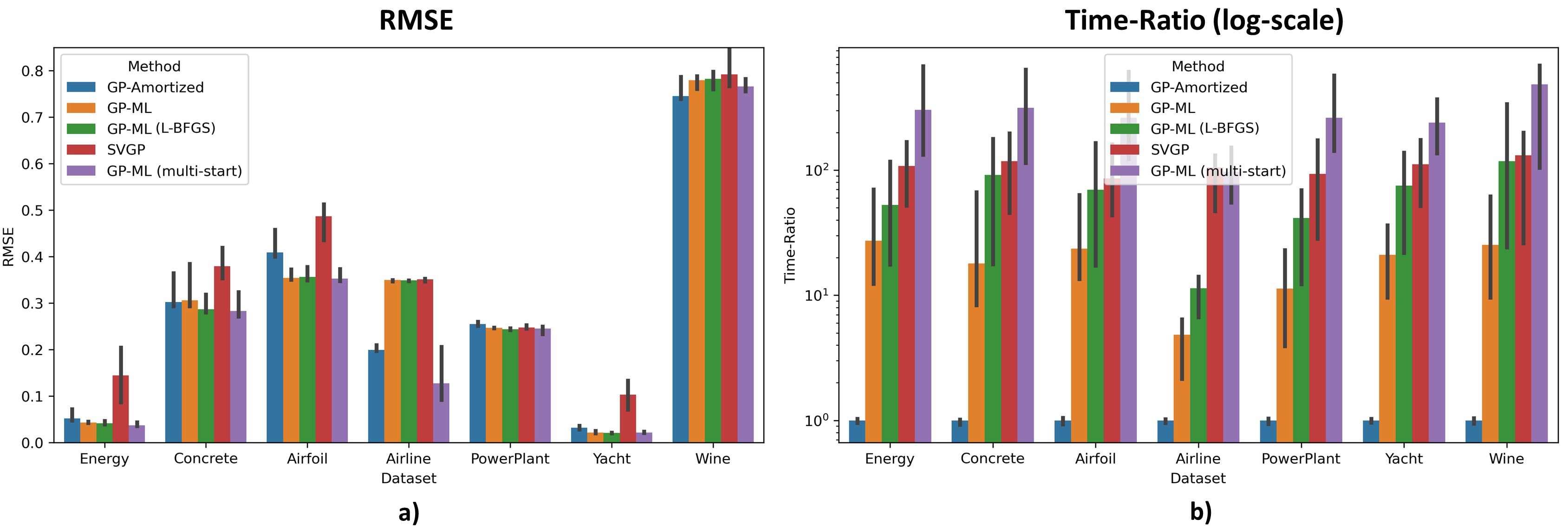}
	\caption{In a) the RMSE scores of each method are shown for held-out test datapoints. For each dataset several kernels $\{\mathcal{S}^{(1)},\dots,\mathcal{S}^{(m)}\}$ are evaluated where each bar shows the median RMSE value and the error-bars show the 20th and 80th percentiles of the RMSE scores of the different kernels. In b) the corresponding ratios of inference times to our method is shown in log-scale.}
	\label{fig:mainresultssingle}
\end{figure*}
\paragraph{Noise variance fine-tuning.}
The noise level is a crucial property of a dataset and determines the predictive performance significantly. We therefore do a dedicated fine-tuning phase after the initial training phase of minimizing the negative marginal-likelihood $\mathcal{L}(\psi,\{(\mathcal{D}_{l},\mathcal{S}_{l})\}_{l=1}^{L})$ in (\ref{main_loss}). We do the fine-tuning via minimizing the extended loss
\begin{align}
\alpha\mathcal{L}(\psi,\{(\mathcal{D}_{l},\mathcal{S}_{l})\}_{l=1}^{L})+\beta \frac{1}{L}\Vert \sigma^{*}_{1:L} - \hat{\sigma}_{1:L} \Vert_{2}^{2},
\end{align}
where we additionally regularize the prediction of the noise-variances $\hat{\sigma}_{1:L}\in\mathbb{R}^{L}_{+}$ to be close to the known ground-truth noise-variances $\sigma^{*}_{1:L}\in\mathbb{R}^{L}_{+}$.  Importantly, we only draw \textit{positive} samples $(\mathcal{D}_{l},\mathcal{S}_{l})$. We call this step a fine-tuning step, as we only do it on significantly fewer datasets than in the first phase. We observe a major increase in robustness of the noise-prediction on real-world datasets. In Appendix B we show the impact of the fine-tuning.

\section{Related Work}
\paragraph{Amortized inference with fixed kernel structure.}
Our method extends the work of \cite{AHGP} to enable amortization over the combined space of datasets and kernel structures. Compared to \cite{AHGP} our network is not restricted to a single kernel, meaning that practitioners can insert any kernel structure $\mathcal{S}$ from our space and can utilize the amortization network out of the box for parameter inference via $\hat{\phi}_{\mathcal{S}}=g_{\psi}(\mathcal{D},\mathcal{S})$. From a software perspective, we can view the work of \cite{AHGP} as an emulation of the inference functionality of a typical GP framework for a fixed kernel through a large neural network. Our method enlarges this emulation further via rendering the kernel configurable directly in the neural network. 

\paragraph{Amortized model selection.} In \citet{KernelIdentWithTrafo} an amortized structure selection is proposed. Here, the kernel structure $\mathcal{S}$ itself is predicted as a sequence of tokens via an amortization network $g(\mathcal{D})$. After selecting $\mathcal{S}$ the hyperparameters of the kernel needs to be trained via maximum likelihood. Our method complements their method, as after taking care that the kernel spaces are identical, one might use our method to predict the kernel parameters of the selected kernel structure. This would amortize the full pipeline of kernel selection and hyperparameter optimization.

\paragraph{Kernel grammar.}
Our input space is based on the kernel grammar in \cite{CKS}. The kernel grammar is part of a greater research line called the \textit{Automatic Statistician} \citep{CKS,AutomaticStatistician,bitzer2022structural}, which tries to infer interpretable GP models and dataset description from data in an automatic way. Our work can be used to enhance the GP parameter inference for each GP representation in the search procedure.

\paragraph{Hypernetworks.}
Our method can be seen as a hypernetwork \citep{hypernetworks} for Gaussian process models. Usually, a hypernetwork predicts neural network weights from some sort of input. The input can be a hyperparameter \citep{hypernetworkHyperparamInp} or a latent representation of a layer \citep{hypernetworks}. Notably, \cite{hypernetworkUnknownArch} proposed a hypernetwork to predict the neural network weights for a fixed dataset with a description of the architecture as input, which can be seen as a related task compared to ours in the neural network world. We note that our method predicts Gaussian process hyperparameters rather than neural network weights and amortizes over the combined space of datasets and kernel descriptions.

\paragraph{Prior-Fitted-Networks.}
In \cite{PFN} a method called \textit{Prior-Fitted-Networks (PFN)} is proposed. Here a transformer is used to form an end-to-end prediction from a dataset $\mathcal{D}$ and a test point $x^{*}$ to a predictive distribution $p(y^{*}|x^{*},\mathcal{D})$ of a given prior. The difference to our approach is two-fold. Firstly, we predict a full GP via inferring its parameters. A PFN predicts only slices of the predictive distribution. Secondly, we render the prior configurable via making the kernel configurable. In this way, practitioners can include prior knowledge of the task at hand.
\begin{table*}[t]
\begin{center}
\caption[RMSE values on real-world datasets (vs AHGP-SE-ARD)]{Average RMSE values over 20 train/test splits for each dataset using our method (GP-Amortized) equipped with an RBF kernel and using AHGP-SE-ARD. Marked values (*) are significantly smaller measured via a two-sample t-test ($\alpha=0.025$).}	
\label{tab:ahgp}
\begin{tabular}{lrrrrrrr}\toprule \textbf{RMSE} &    Energy &  Concrete &  Airfoil &  Airline &  PowerPlant &     Yacht &      Wine \\\midrule 
	GP-Amortized &  0.0830 &  0.3635 &  *0.4484 &  0.2866 &    0.2467 &  0.0606 &  ~0.9287 \\
	AHGP-SE-ARD &  0.0800 &  0.3755 &  ~0.5965 &  0.2722 &    0.2459 &  0.0602 &  *0.7962 \\\bottomrule
	\end{tabular}
\end{center}	
	
\end{table*}
\section{Experiments}
In the following section, we empirically analyze our amortization scheme on regression benchmarks and compare against common methods to do GP inference. First, we illustrate the prediction capabilities on toy datasets. Subsequently, we analyze the learning behavior and the performance on real-world datasets. In the last subsection, we analyze ensembling as a possible plug-and-play extension of our method.

\paragraph{Experimental setting.}
We train our amortization network on a stream of mini-batches of size $L=128$ for a total of 9 million dataset-kernel pairs $(\mathcal{D}_{l},\mathcal{S}_{l})$ in the initial training phase. We continue the training with the noise-variance fine-tuning phase, which is performed over 200.000 dataset-kernel pairs. In both phases, we utilize $\mathrm{SE},\mathrm{LIN}$ and $\mathrm{PER}$ and its 2-gram multiplications like, e.g. $\mathrm{SE}\times\mathrm{LIN}$ as base-symbols and simulate datasets of size $n\in[10,250]$ and input dimension $d\in[1,8]$. Further training details can be found in Appendix A. 

\paragraph{Performance on simulated data.}
We evaluate the final amortization network $g_{\psi}$ via its inference capabilities on unseen datasets $\tilde{\mathcal{D}}$. Each unseen dataset $\tilde{\mathcal{D}}$ is splitted into training $\tilde{\mathcal{D}}_{train}$ and test dataset $\tilde{\mathcal{D}}_{test}$ and we evaluate for a given kernel expression $\mathcal{S}$ the predictive performance of the final GP with predicted hyperparameters $\hat{\phi}_{\mathcal{S}}=g_{\psi}(\tilde{\mathcal{D}}_{train},\mathcal{S})$ on $\tilde{\mathcal{D}}_{test}$. We give example predictive distributions on simulated datasets in Figure \ref{fig:toydata} a). We observe that our amortization network leads to accurate predictive distributions - notably only via evaluating a neural network to predict the kernel parameters. We add several more prediction plots in Appendix B, including plots with misspecified kernels, small datasets and more complex kernels.

Furthermore, we show test-RMSE and test-NLL scores of 200 unseen, simulated dataset-kernel pairs $(\tilde{\mathcal{D}}_{l},\mathcal{S}_{l})$ in Figure \ref{fig:toydata} b) for our method and for a GP with Type-2-ML hyperparameter inference. The datasets are sampled from the same distribution as used for the initial training phase. We observe that our method leads to very similar RMSE and NLL scores compared to Type-2-ML inference. This illustrates the quality of the predicted hyperparameters. In Appendix B, we analyze the predictive performance of both approaches for different number of training datapoints.
\paragraph{Regression benchmarks.}
Our main evaluation considers seven real-world datasets. We split each dataset into a training and test set (we set $\mathrm{n_{train}}=500$ for all datasets except the smaller datasets $\mathrm{Airline}$ and $\mathrm{Yacht}$, details in Appendix A) and evaluate the predictive performance over a set of $m=24$ kernels $\{\mathcal{S}^{(1)},\dots,\mathcal{S}^{(m)}\}$ that are drawn randomly. We compare against the standard way of GP hyperparameter inference via Type-2-ML. We consider three versions of Type-2-ML two with a single run from initial parameters where we optimize via Adam [GP-ML] and via L-BFGS [GP-ML (L-BFGS)] and one with 10 randomized restarts optimized with Adam [GP-ML (multi-start)]. For GP-ML and GP-ML (multi-start), we maximize the marginal likelihood via \textrm{Adam} with $\mathrm{lr}=0.1$ for 150 iterations and early-stop once the loss is converged. Furthermore, we compare against Sparse-Variational GP (SVGP) \cite{svgp}, where we use $I=0.5n$ inducing points. For more details see Appendix A.

 In Figure \ref{fig:mainresultssingle} a), we show the resulting test-RMSE scores of the respective method. Here, the black bars correspond to percentiles of the RMSE scores of the set of different kernels $\{\mathcal{S}^{(i)}\}_{i=1}^{m}$. We observe that our methods leads to comparable predictive performance to  Type-2-ML. Importantly, we observe in Figure \ref{fig:mainresultssingle} b) that our proposed method leads to a \textit{major decrease in inference time for a diverse set of kernels $\{\mathcal{S}^{(i)}\}_{i=1}^{m}$}. This goes so far that for certain kernel structures our method is 800 times faster than Type-2-ML with restart. We show NLL scores in Appendix A.
\begin{figure*}[t]
	\centering
	\includegraphics[width=0.99\linewidth]{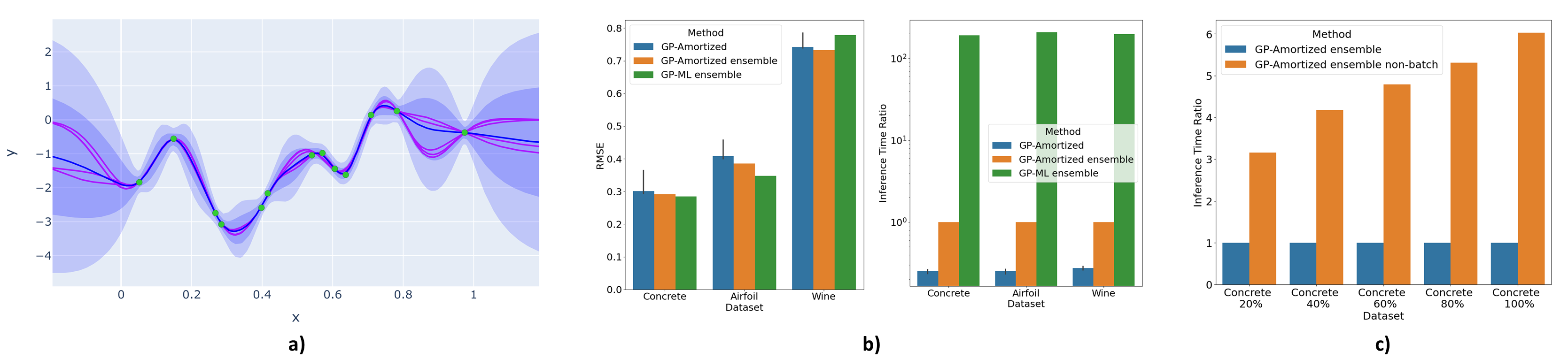}
	\caption{Fast ensembling. a) Predictive distribution resulting from a fast ensemble of five kernel structures (in blue). In red, we show the predictive means of the single GP's. b) RMSE and inference time ratios for the fast ensembling based on our amortization model (error bars of standard amortization are over the same 24 kernels that are in the ensemble). c) Ratios of inference time between batch and non-batch ensembling shown over different dataset sizes.}
	\label{fig:ensembleplots}
\end{figure*}
\paragraph{Comparison to AHGP.} We further compare against AHGP on a fixed kernel. We note that for any new kernel structure the method of \cite{AHGP} needs to be redesigned and retrained - rendering it less flexible than our method. We investigate the performance differences on the ARD-RBF kernel, which is part of our considered kernel space. Here, we use the adapted version of the AHGP architecture to the RBF kernel presented in \cite{rehn2022amortized} (AHGP-SE-ARD). We trained AHGP-SE-ARD on the same data distribution as our method for 9 million datasets and configured the architecture to have approximately the same capacity (see Appendix A for details). For evaluation/inference on a dataset $\tilde{\mathcal{D}}$ we equip our method with the ARD-RBF kernel as input, thus $\hat{\phi}=g_{\psi}(\tilde{\mathcal{D},}\prod_{i=1}^{d}\textrm{SE}_{i})$. We show mean test-RMSE scores on the real-world datasets for both methods in Table \ref{tab:ahgp}. It can be seen that on five out of seven datasets the performance of both method is very similar - indicating that amortization over kernel structures does not induce significant performance drops compared to using a fixed kernel. On \textit{Airfoil} our method is significantly better and on \textit{Wine} the AHGP-SE-ARD method. We think that this difference might eventually vanish with more datasets in the training phases.

\paragraph{Fast ensembling.} Our method offers a general inference machine for GP hyperparameters over a structured kernel space. This can be utilized to construct ensembles in a computationally efficient way. We construct a Bayesian-Model-Average (BMA) over a set of kernel structures $\{\mathcal{S}^{(1)},\dots,\mathcal{S}^{(m)}\}$ , where we use the predicted marginal likelihood values as ensemble weights (see Appendix A). In \ref{fig:ensembleplots} a) we show an example of the fast ensembling over five kernel structures. Importantly, we observe a high diversity in the predictive mean functions, which is a desirable property for an ensemble. In \ref{fig:ensembleplots} b) we show the predictive performance of the fast ensemble over the set of 24 kernels and compare it to the range of predictive scores of the non-ensemble predictions (we show the three datasets that had the biggest diversity of RMSE scores over kernels). Furthermore, we compare against an ensemble, with Type-2-ML inferred GP parameters. Firstly, we observe that ensembling results in the expected performance gain over standard predictions and secondly, we see that our method offers a drastic speed-up compared to the conventional method. 

Importantly, our architecture is particularly tangled to process this kind of ensembles over kernel structures, as it only needs to process the dataset-encoder once and can process multiple kernels in a batch through the kernel-encoder-decoder. This leads to an additional speed-up for inferring ensembles that can be seen in Figure  \ref{fig:ensembleplots} c).

\paragraph{Limitations.}
We note that predicting kernel parameters instead of optimizing them also has limitations. In particular, our method tends to favor conservative predictive distributions with broader prediction intervals out-of-data (see Appendix B). While this is often beneficial, it might not be desired, for example for extrapolation tasks with the periodic kernel where our method favors explaining the data via the lengthscale of the periodic kernel rather than the periodicity (see qualitative analysis in Appendix B).

\section{Conclusion}
In this work, we proposed an amortization scheme for the hyperparameters of Gaussian process models. The main novelty is to amortize over the combined space of datasets and kernel structures. Our proposed amortization network is explicitly designed to cope with the respective symmetries of this task. In our experiments, we show a drastic speed-up in inference time for diverse kernel structures. At the same time, we show that our method can predict kernel parameters that lead to competitive predictions on real world data.

\bibliography{amor_struct_gp}
\title{Amortized Inference for Gaussian Process Hyperparameters of Structured Kernels\\(Supplementary Material)}
\onecolumn 
\maketitle
\appendix
In the following sections, we give further information about our method. We start by giving more experimental details. Then we illustrate the application of our method on an extended set of simulated datasets via plots of the predictive distributions. Finally, we give the proofs for the invariances and equivariances of our amortization network.
\section{Experimental Details}
\paragraph{Architecture - Dataset-Encoder.} The dataset-encoder consists of four transformer blocks each consisting of a stack of Transformer-Encoder sublayers \citep{AttentionIsAllYouNeed} where each sublayer has a multi-head-self-attention layer and an element-wise MLP layer as the two trainable parts. We don't use dropout and positional encodings in our architecture as it would disable equivariances \citep{SetTransformer}.
\begin{table}[h]
\centering
\caption{Dataset-Encoder Configuration.}
	\begin{tabular}{llll}\hline
		\textbf{Block} & Num. of Layers & Embedding Dim. & Hidden Dim. in MLP \\\hline
		\textbf{Transformer} no. 1 (used in step 3) & 4                & 256                 & 512              \\
		\textbf{Transformer} no. 2 (used in step 5) & 4                & 256                 & 512           \\
		\textbf{Transformer} no. 3 (used in step 7) & 4                & 512                 & 512     \\
		\textbf{Transformer} no. 4 (used in step 9) & 4                & 512                 & 512 \\  
		\hline    
	\end{tabular}
\end{table}
\paragraph{Architecture - Kernel-Encoder-Decoder.} The kernel-encoder-decoder consists of two stacks of $\textbf{Kernel-Encoder-Block}$ layers, which are also specified via an embedding dimension and the hidden dimension of its MLP layer. Furthermore, the kernel-encoder-decoder also contains one transformer block.
\begin{table}[h]
\centering
	\caption{Kernel-Encoder-Decoder Configuration.}
	\begin{tabular}{llll}\hline
		\textbf{Block} & Num. of Layers & Embedding Dim. & Hidden Dim. in MLP \\\hline
		$\textbf{Kernel-Encoder-Block}$ stack 1 (used in step 1) & 3                & 512               & 1024              \\
		\textbf{Transformer} block (used in step 2) & 4                & 512               & 1024           \\
		$\textbf{Kernel-Encoder-Block}$ stack 2 (used in step 4) & 3                & 512                 & 1024   \\ 
		\hline    
	\end{tabular}
\end{table}
\paragraph{Architecture - Output layer.} Each base-symbol specific \textbf{MLP} layer has one hidden layer with dimension $d_{h}=200$. The \textbf{MLP} layer for the noise variance prediction consists of two hidden layers with dimension  $d_{h_{1}}=200$ and $d_{h_{2}}=100$.

\paragraph{Sampling distribution.} We use as base symbols $\mathrm{SE},\mathrm{LIN}$ and $\mathrm{PER}$ and its two gram multiplications like, e.g. $\mathrm{SE}\times\mathrm{LIN}$. We simulate datasets of sizes between $n=10$ and $n=250$ (drawn uniformly) and input dimensions between $d=1$ and $d=8$. Here, we prefer smaller dimensions via sampling $d$ from a geometric distribution with $p=0.25$ and clip them to $d\le8$. We sample the number of addends in each subexpression $\mathcal{S}_{i}$ also from a geometric distribution with $p=0.6$. Each dataset is generated via sampling a kernel structure, sampling the input dataset uniformly from $[0,1]^{d}$, then sampling the kernel parameters and noise variance from their prior. Finally, we sample the output of the dataset from the respective GP at the respective input locations. We use the following base kernel parameterizations,
\begin{align*} 
k_{\mathrm{SE}}(x,x')=\sigma_{k}^{2}\mathrm{exp}\bigg(-\frac{1}{2} \frac{(x-x')^{2}}{l^{2}} \bigg),~~k_{\mathrm{PER}}(x,x')=\sigma_{k}^{2}\mathrm{exp}\bigg(-\frac{1}{2} \frac{\mathrm{sin}^{2}(\pi|x-x'|/p)}{l^{2}} \bigg),~~k_{\mathrm{LIN}}(x,x')=\sigma_{k}^{2}x\,y + \sigma_{c}^{2}
\end{align*}
with priors, $\sigma_{k}^{2} \sim \mathrm{Gamma}(2.0,3.0)$, $\sigma_{c}^{2} \sim \mathrm{Gamma}(2.0,3.0)$, $p \sim \mathrm{Gamma}(2.0,3.0)$ and $l \sim \mathrm{Gamma}(2.0,5.0)$. For the noise variance we use the prior $\sigma^{2}\sim\mathrm{Exp}(\frac{1}{0.15^{2}})$.
\begin{figure}[t]
	\centering
	\includegraphics[width=1.0\linewidth]{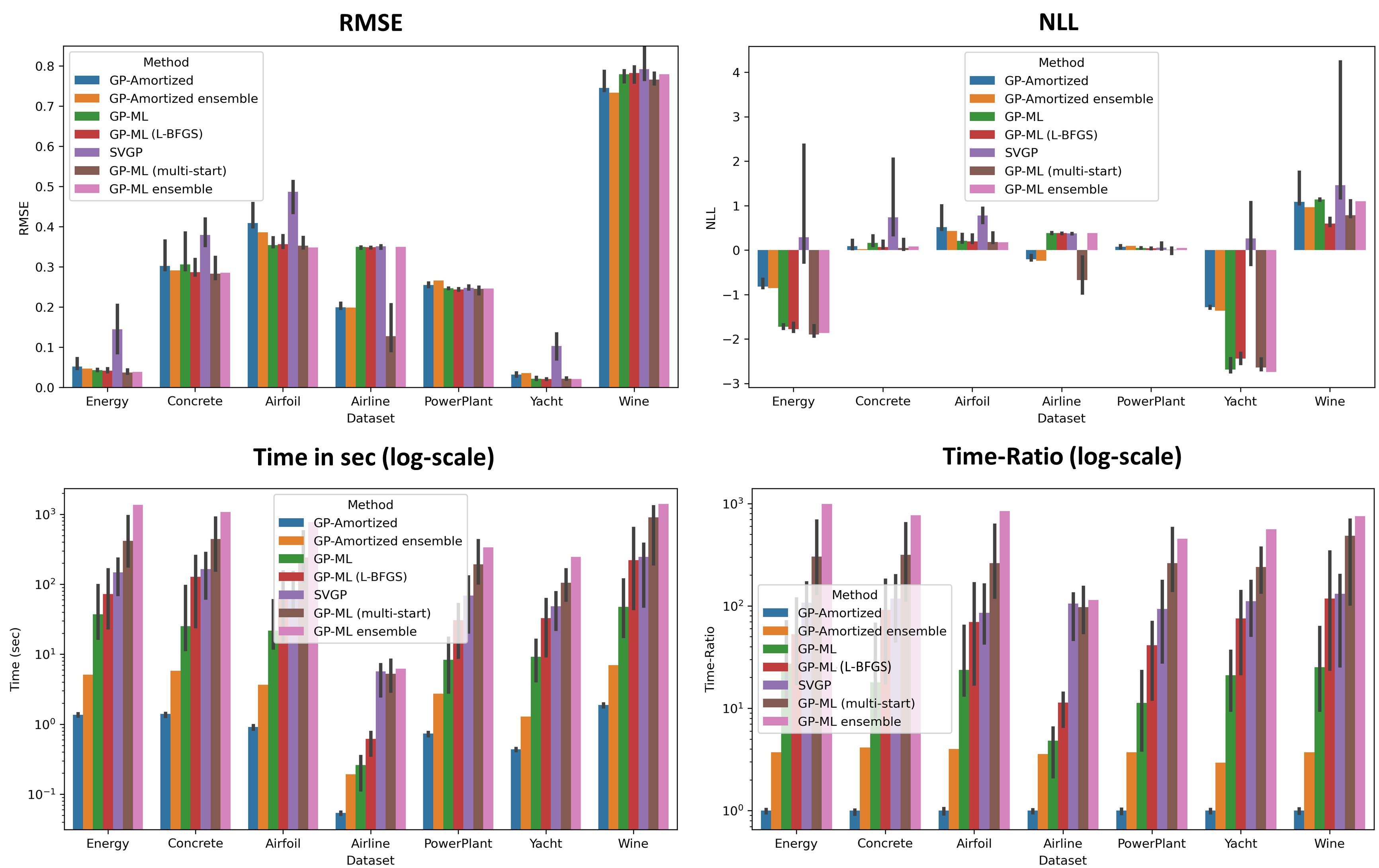}
	\caption{In the top-left plot the RMSE scores of all considered methods are shown evaluated on the respective test set.  Several kernels $\{\mathcal{S}^{(1)},\dots,\mathcal{S}^{(m)}\}$ are evaluated where each bar shows the median RMSE value and the error-bars show the 20th and 80th percentiles of the RMSE scores of the different kernels. In the top-right plot, the NLL scores are shown and in the bottom-right plot the corresponding ratios of inference times to our method are shown in log-scale. In the bottom-left plot, the absolute inference times in seconds are shown for the corresponding method.}
	\label{fig:predictionplotmainresultscomplete}
\end{figure}

\paragraph{Training - Phase 1 and 2.} In the first phase, we train the amortization network via a batch-size of $B=128$ for a total of $9$ million datasets, where in each batch we simulate the datasets on-the-fly. Thus, each dataset in the training phase only appears once. We use RAdam as optimizer with a constant learning rate of $\mathrm{lr}=\mathrm{2\times 10^{-5}}$. In the second phase the extended loss is configured with parameters $\alpha=10$ and $\beta=1$. Besides that, we use the same configuration as in the first phase. We train in this phase on 200 thousand datasets.

\paragraph{Datasets.} All datasets are publically available (\textrm{Airline} can be accessed for example via  \url{https://kaggle.com}. The other datasets are UCI datasets from \url{https://archive.ics.uci.edu/ml/datasets.php}). We use for all datasets $n_\textrm{train}=500$ except the smaller datasets \textrm{Airline} and \textrm{Yacht} for which we use 100 and 250 training datapoints. The training datapoints are drawn uniformly from the complete dataset. We use the rest of the dataset as test-set, clipped to a maximum of $n_\textrm{test}=400$. 

\paragraph{Type-2-ML GP settings.}  In our main evaluation, we consider three competitor approaches for full GP's with Type-2-ML hyperparameter inference: Firstly, GP-ML and GP-ML (multi-start) where we do Type-2-ML optimization via Adam with one and with multiple restarts. Here, we use the following setting. We use the Adam optimizer with a learning rate of 0.1 for 150 steps to optimize the marginal likelihood, where we early-stop the optimization once the marginal-likelihood has converged. We found that this setting provides a good trade-off between accuracy and inference time. For the single-run version we use the same initial parameter values for all datasets, where we set all initial parameter values of the kernels to one, e.g. $\sigma_{k}=1.0$ and the initial likelihood noise to $\sigma=0.2$. For the multi-start version, we do 10 restarts, where in each restart the initial parameter values are sampled from the respective prior. The methods are based on GPyTorch \citep{gardner2018gpytorch}. Furthermore, we consider optimization of the marginal likelihood via L-BFGS, to which we refer to as GP-ML (L-BFGS). Here, we use GPflow \citep{GPflow2017} as implementation and use the default optimization parameters of the library.

\paragraph{Sparse-Variational GP settings.} For the Sparse-Variational GP (SVGP) we used $I=0.5n$ number of inducing points, where $n$ is the number of training points in the dataset. Furthermore, we use Adam to optimize the ELBO with respect to the inducing point locations and GP parameters. Here, we trained for 500 iterations (full-batch) with a learning rate of $lr=0.03$, with early stopping once the ELBO converged. 

\paragraph{Evaluation set of kernels.} In the main evaluation and for fast-ensembling we utilize a set of 24 kernel structures $\{\mathcal{S}^{(1)},\dots,\mathcal{S}^{(m)}\}$. Here, we use in each input dimension the same substructure, thus $\mathcal{S}_{i}^{(l)}=\mathcal{S}_{j}^{(l)}$ for $j\neq i$ and all $l=1,\dots,m$ and generate the substructures in the following way. We include the set of base-symbols $\{\mathrm{SE},\mathrm{PER},\mathrm{LIN},\mathrm{SE}\times\mathrm{PER},\mathrm{SE}\times\mathrm{LIN},\mathrm{LIN}\times\mathrm{PER}\}$. For example $\mathcal{S}^{(1)}$ with $\mathcal{S}^{(1)}_{i}=\mathrm{SE}_{i},i=1,\dots,d$ and $\mathcal{S}^{(2)}$ with $\mathcal{S}^{(2)}_{i}=\mathrm{PER}_{i},i=1,\dots,d$, are part of the set of 24 kernel structures. Furthermore, we sample six pairs, e.g. $\mathcal{S}_{i}=\mathrm{SE}_{i}+\mathrm{LIN}_{i},i=1,\dots,d$, six triples and six quadruples from the set of base symbols each without replacement to generate the 24 kernel structures. In this way, we apply the same set of kernel (sub-)structures on each dataset - only the number of input dimension changes.

\paragraph{AHGP-SE-ARD settings.} For the AHGP-SE-ARD model, we replicated the architecture of \citet{rehn2022amortized}, which uses the dataset-encoder provided by \cite{AHGP} and a custom head to deal with the RBF kernel. We equip the dataset-encoder with 14 \textit{LocalTransformer} blocks and 14 \textit{GlobalTransformer} blocks (see \cite{AHGP}) each specified with an embedding size of 512. Furthermore, we use for the lengthscale and variance MLP's (see \citet{rehn2022amortized}) two hidden layers with 1024 and 512 nodes. We train the network on the same data distribution as our method for 9 million datasets, where we used RAdam as optimizer with a learning rate of $\mathrm{lr}=\mathrm{2\times 10^{-5}}$ and a batch size of $L=32$.

\paragraph{Fast-ensembling details.}
For our fast-ensembling approach, we use a set of kernel structures $\{\mathcal{S}^{(1)},\dots,\mathcal{S}^{(m)}\}$ and define the resulting predictive distribution of our ensemble model for a dataset $\mathcal{D}$ with input $\mathbf{X}$ and output $\mathbf{y}$ via
\begin{align*}
p(y^{*}|x^{*},\mathcal{D}):=\frac{1}{m}\sum_{l=1}^{m} w_{l} \,p\bigg(y^{*}\bigg|x^{*},\mathcal{D},\phi_{\mathcal{S}^{(l)}}=g_{\psi}(\mathcal{D},\mathcal{S}^{(l)})\bigg),
\end{align*}
where $p\bigg(y^{*}\bigg|x^{*},\mathcal{D},\phi_{\mathcal{S}^{(l)}}=g_{\psi}(\mathcal{D},\mathcal{S}^{(l)})\bigg)$ is the predictive distribution of the GP with kernel structure $\mathcal{S}^{(l)}$ and predicted parameters $\phi_{\mathcal{S}^{(l)}}=g_{\psi}(\mathcal{D},\mathcal{S}^{(l)})$
and where $w_{l} =\frac{\tilde{w}_{l}}{\sum_{l=1}^{m}\tilde{w}_{l}}$ with $\tilde{w}_{l}:=p(\mathbf{y}|\mathbf{X},\phi_{\mathcal{S}^{(l)}}=g_{\psi}(\mathcal{D},\mathcal{S}^{(l)}))$ is the (predicted) marginal likelihood value for kernel structure $\mathcal{S}^{(l)}$.

\paragraph{Comparison of all considered models.} In Figure \ref{fig:predictionplotmainresultscomplete} the evaluation plots are shown for all considered methods. Here, we also show NLL scores. The biggest difference in NLL compared to the full-GP Type-2-ML methods is present for datasets where the predictions are already very precise (measured via RMSE) like on Energy or Yacht aka, where the NLL scores are already very small. It is worth noting that we outperform SVGP in terms of NLL on all datasets except Powerplant.

\section{Further Experiments}

\begin{figure}[h]
	\centering
	\includegraphics[width=0.795\linewidth]{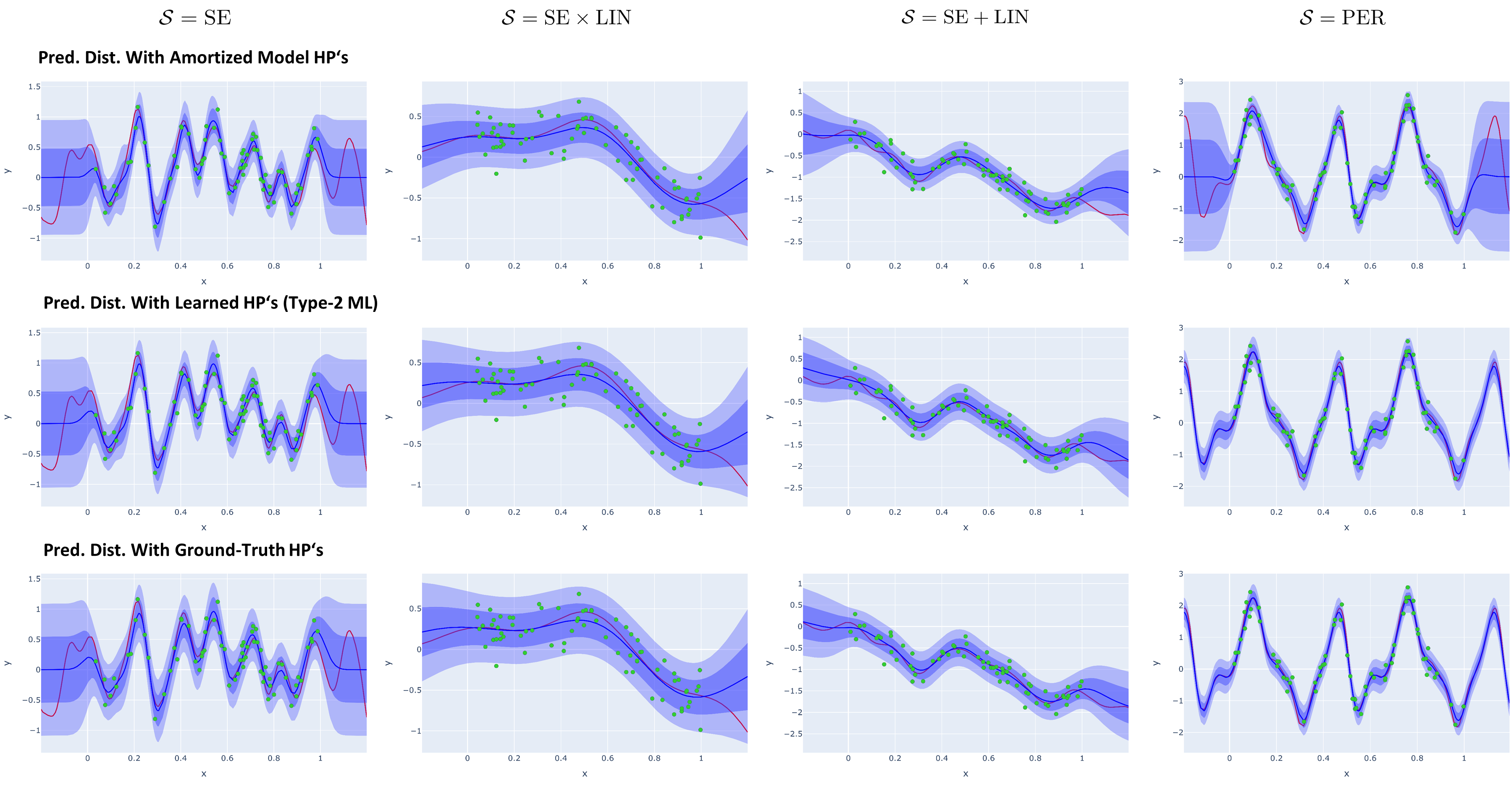}
	\caption[Predictions - Correct Kernel]{We show predictive distributions with kernel parameters of our method (top), Type-2-ML (middle) and with ground truth hyperparameters (bottom), where in each column a different input kernel structure $\mathcal{S}$ is used and  $\mathcal{S}=\mathcal{S}_{gt}$.}
	\label{fig:predictionplothighnoise}
\end{figure}

\paragraph{Qualitative Analysis of Predictive Distributions.} We do a qualitative analysis of the predictive distributions for 1D datasets for different settings. In the plots (see for example Figure \ref{fig:predictionplothighnoise}) datasets $\mathcal{D}^{*}$ are sampled from a GP with ground truth kernel $\mathcal{S}_{gt}$. We compare the predictive distributions induced by our predicted GP parameters $\phi^{*}=g_{\psi}(\mathcal{D}^{*},\mathcal{S})$ to the predictive distributions induced by the learned GP parameters $\phi_{ML}$ via Type-2-ML and against the predictive distributions with the ground truth hyperparameters $\phi_{gt}$. 

The first setting (Figure \ref{fig:predictionplothighnoise}) considers a \textit{correctly specified kernel structure}, where the input kernel structure $\mathcal{S}$ is the same as the ground truth kernel structure $\mathcal{S}_{gt}$. In Figure \ref{fig:predictionplothighnoise}, the predictive distributions of the three methods can be seen for four different kernel structures. 
We see that the noise range is captured well by our method, as well as the in-distribution data-fit. In the extrapolation regime left and right of the data, the confidence intervals of our method appear reasonable for $\mathrm{SE}$, $\mathrm{SE}\times \mathrm{LIN}$ and $\mathrm{SE}+ \mathrm{LIN}$. However, the method makes a conservative prediction for the $\mathrm{PER}$ kernel via ignoring the periodicity in the data, whereas the type-2 ML and ground truth kernel extrapolate less conservative, which in this case leads to accurate predictions.

\begin{figure}[h]
	\centering
	\includegraphics[width=0.795\linewidth]{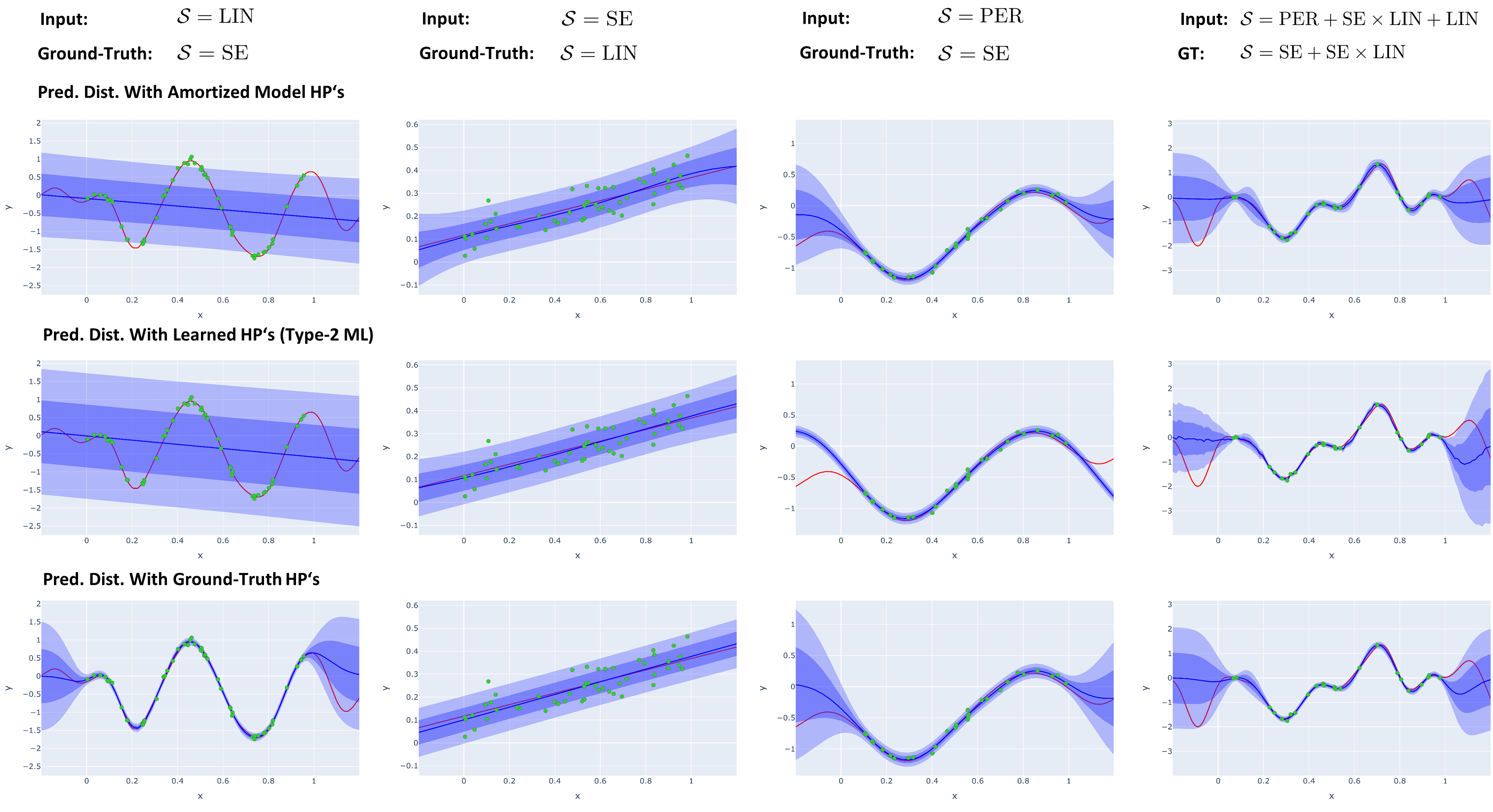}
	\caption[Predictions - Misspecified Kernel]{We show the predictive distributions with kernel hyperparameters predicted via our method (top), learned via Type-2-ML (middle) and with ground truth hyperparameters (bottom), where in each column a different input kernel $\mathcal{S}$ and a different ground-truth kernel $\mathcal{S}_{gt}$ is used and  $\mathcal{S}\neq\mathcal{S}_{gt}$. In red, the underlying ground-truth function is shown.}
	\label{fig:predictionplotmisspecified}
\end{figure}

For the second setting (Figure \ref{fig:predictionplotmisspecified}) a \textit{misspecified kernel structure} is used, where the input kernel structure $\mathcal{S}$ is not the same as the ground truth kernel structure $\mathcal{S}_{gt}$. We see for example for the dataset coming from the linear kernel (second column from left) that the predicted lengthscale of the specified $\mathrm{SE}$ kernel appears to be sufficiently large. When the linear kernel is applied to a dataset coming from an $\mathrm{SE}$ kernel (first column from left)
 we see that the noise variance prediction is higher than the ground-truth noise (which was $\sigma=0.05$), which is desirable for the linear kernel as the GP needs to explain the data with noise in this case. 
Arguably, the most interesting case is, when a periodic kernel $\mathrm{PER}$ is specified as input and the input data is coming from an $\mathrm{SE}$ kernel (third column from left). Here, we see that the predictive-distribution of the Type-2-ML estimate extrapolates aggressively and overfits to the data (we made sure to find a very high marginal likelihood value here via repeating the optimization 30 times). In this case, the conservative prediction of our method for the periodic kernel is beneficial, as it leads to reasonable prediction intervals. As the last kernel pair, we use two more complex kernels as ground truth and input kernel. Here, we observe that our method can also deal with more complex kernels and leads to a reasonable predictive distribution.  

\begin{figure}[h]
	\centering
	\includegraphics[width=0.79\linewidth]{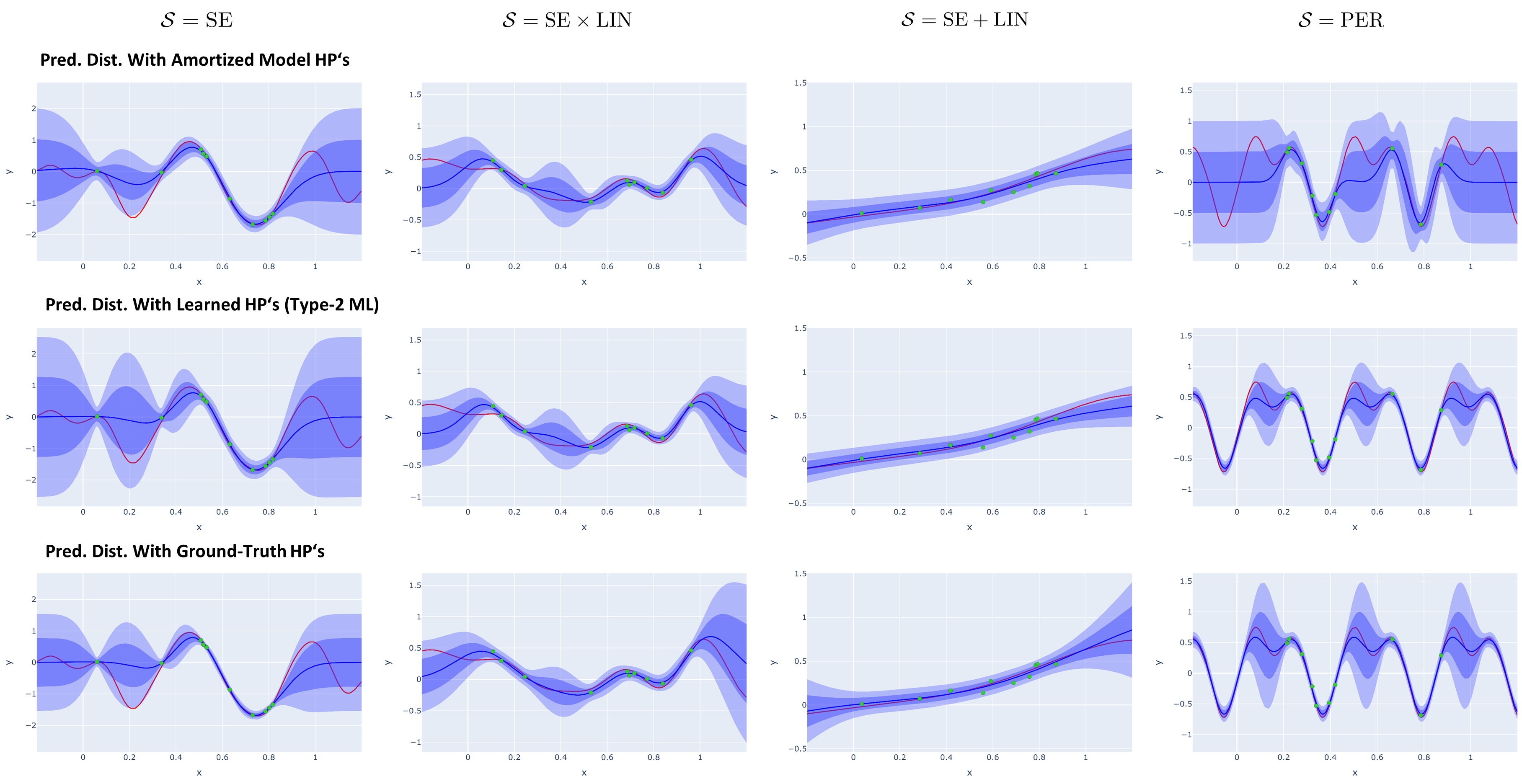}
	\caption[Predictions - Smaller Datasets]{We show predictive distributions with kernel hyperparameters of our method (top), Type-2-ML (middle) and with ground truth hyperparameters (bottom), where in each column a different input kernel structure $\mathcal{S}$ is used and  $\mathcal{S}=\mathcal{S}_{gt}$. In red we show the underlying ground-truth function. Here, we consider small datasets with $n=10$.}
	\label{fig:predictionplotsmalldataset}
\end{figure}

\begin{figure}[h]
	\centering
	\includegraphics[width=0.70\linewidth]{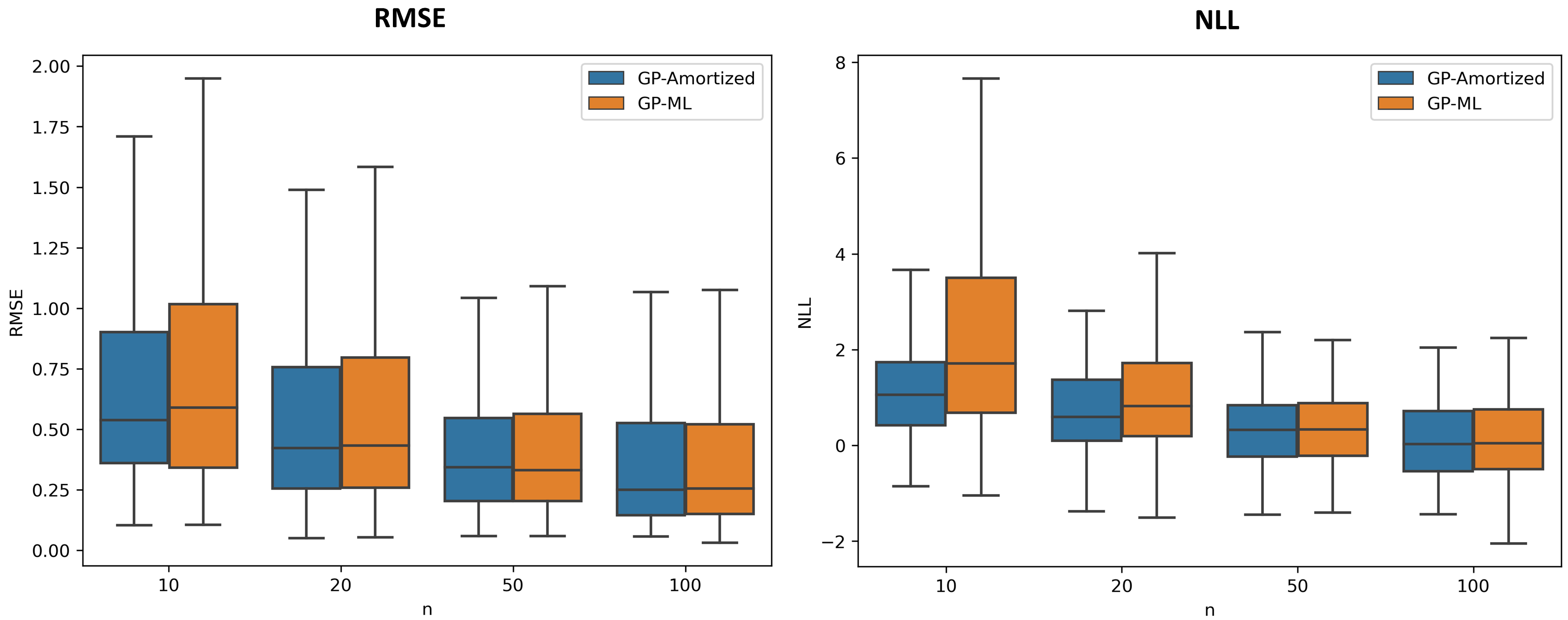}
	\caption[RMSE and NLL for simulated data with varying n]{ We show RMSE and NLL boxplots on simulated data (splitted into train/test) with $d=2$ and varying $n_{train}$. Each sample is a dataset kernel pair $(\mathcal{D},\mathcal{S})$ drawn from our training sampling distribution (except that $d$ and $n$ are fixed). We compare our method (GP-Amortized) against Type-2-ML (GP-ML).}
	\label{fig:rmse_nll_smaller}
\end{figure}

\begin{figure}[t]
	\centering
	\includegraphics[width=0.80\linewidth]{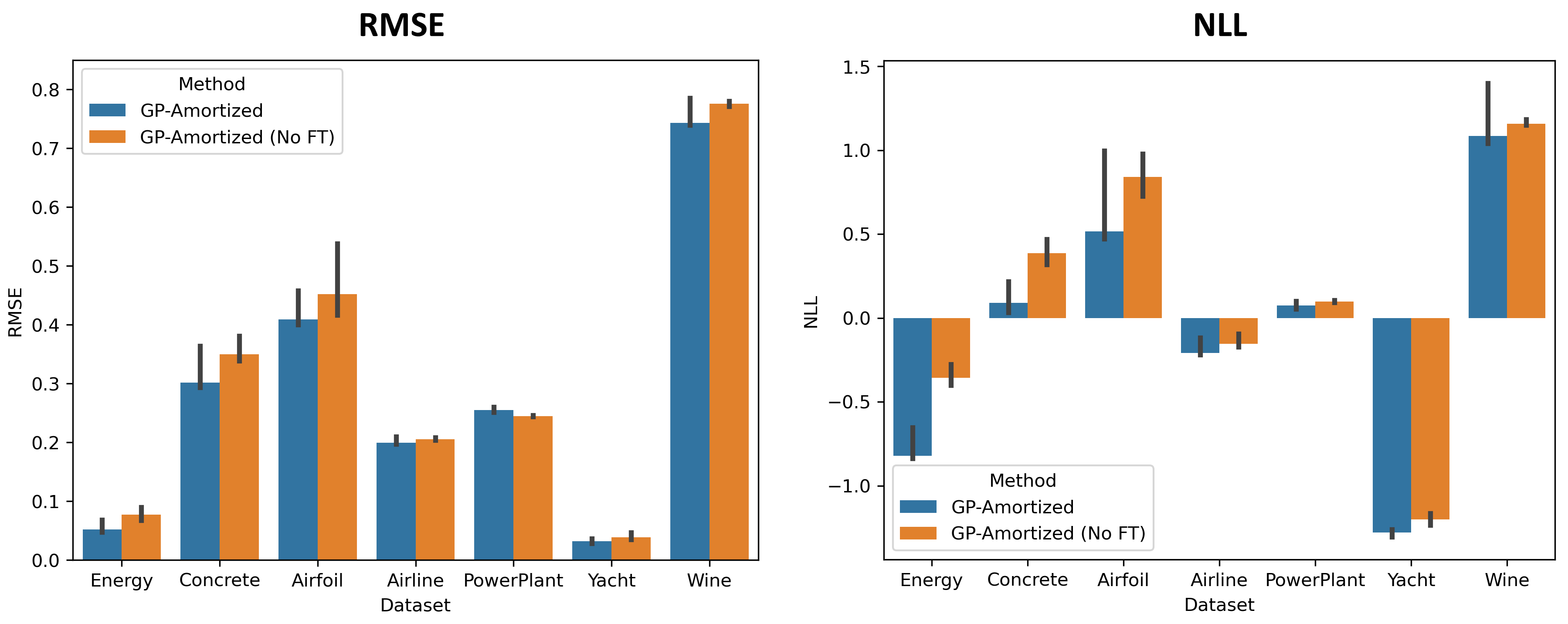}
	\caption[RMSE and NLL fine-tuning]{ RMSE and NLL over kernel structures (error bars) and datasets for our method with noise-variance fine-tuning (GP-Amortized) and without fine-tuning [GP-Amortized (No FT)].}
	\label{fig:noise_ft}
\end{figure}

\begin{figure}[b]
	\centering
	\includegraphics[width=0.99\linewidth]{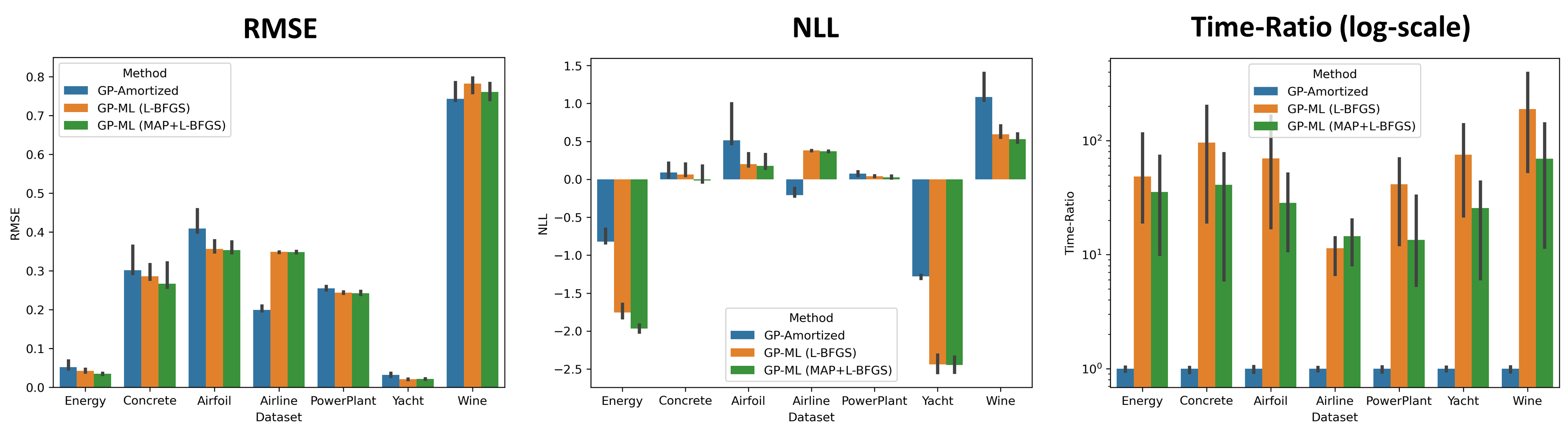}
	\caption[RMSE and NLL fine-tuning]{ RMSE, NLL and time ratios over kernel structures (error bars) and datasets for our method, a GP trained with L-BFGS and a GP with the same priors on the kernel parameters as used for training data generation of the amortization neural network (here also L-BFGS was used).}
	\label{fig:map_comp}
\end{figure}

In the last setting (Figure \ref{fig:predictionplotsmalldataset}), we consider $\textit{smaller datasets}$ with $n=10$. Here, we again use correctly specified kernels. We observe for all kernel structures reasonable prediction intervals and out-of-data predictions. 
We observe for example for the $\mathrm{SE}+\mathrm{LIN}$ that the method recognizes that the data is very linear already with ten datapoints. Notably, we see again a conservative prediction for the periodic kernel and a less conservative, but in this case accurate extrapolation for the Type-2-ML estimates.
\paragraph{Simulation analysis for smaller datasets on inference time.} In Figure \ref{fig:rmse_nll_smaller} we show prediction results on simulated data, with varying sizes of the training set $\mathcal{D}_{train}$ that is used for one-shot prediction of the kernel parameters. 
For each boxplot, we simulate 200 dataset-kernel pairs $(\mathcal{D},\mathcal{S})$  from our sampling distribution, where we fix $d=2$ and vary the number of training datapoints $n_{train}=10,20,50,100$ in order to investigate the impact that the number of training inputs has on the performance difference of our method (GP-Amortized) against Type-2-ML (GP-ML).
 Similar to \cite{AHGP}, we find that in particular for smaller datasets amortization provides more robust kernel parameters, as can be seen for example in terms of improved NLL scores on smaller datasets. Thus, amortizing kernel parameter inference might provide regularization against overfitting of the kernel parameters and thus might be in particular useful in the small-data regime.
\paragraph{Effect of noise variance fine-tuning.}
We investigate the effect of noise variance fine-tuning on the real-world performance in Figure \ref{fig:noise_ft} and observe that on almost all datasets the fine-tuning has a beneficial effect on (median) RMSE and NLL scores. We think that this is the case because the marginal-likelihood landscape is multi-modal with respect to the noise value (high-noise vs low-noise explanation of the data) and the noise variance fine-tuning renders the loss landscape more well-behaved, similar to putting a prior on the GP parameters when doing maximum-a-posteriori (MAP) estimation.
\paragraph{Comparison to MAP estimation.} Using simulated data from a hierarchical GP model with priors on the kernel hyperparameters encodes the prior to a certain extent into the amortization neural network, via implicitly encoding the range of parameter values in the datasets that are used in the training of the amortization network. Thus, one might compare our method against MAP estimation of the kernel parameters using the same priors. In Figure \ref{fig:map_comp} we show a comparison to MAP estimation of the GP hyperparameters with standard ML estimation and with our method. 
We observe that MAP estimation provides a small benefit compared to marginal-likelihood maximization in terms of RMSE and NLL and also in terms of computation time - likely because of a more well-behaved loss landscape, where L-BFGS converges with fewer iterations. However, we note that our method is still orders of magnitudes faster.

\section{Theory}
Before starting to prove the invariance/equivariances in our network, we provide a high-level example on which invariances/equivariances are present in our architecture and why they are important. Our network predicts for a given dataset $\mathcal{D}$ and a given kernel structure $\mathcal{S}$ the kernel parameters and likelihood noise $(\hat{\theta}_{\mathcal{S}},\hat{\sigma}^{2})=g_{\psi}(\mathcal{D},\mathcal{S})$. Firstly, we note that a reshuffling of the order of the dataset elements inside $\mathcal{D}$ should not change the output of $g_{\psi}$, since a reordering of the dataset also does not change the Type-2-ML maximization in the standard GP optimization. Therefore, it should hold for a shuffled dataset $\mathcal{D}_{\pi}$ that
\begin{align*}
(\hat{\theta}_{\mathcal{S}},\hat{\sigma}^{2})=g_{\psi}(\mathcal{D},\mathcal{S})=g_{\psi}(\mathcal{D}_{\pi},\mathcal{S})=(\hat{\theta}_{\mathcal{S},\pi},\hat{\sigma}^{2}_{\pi})
\end{align*}
Furthermore, we consider kernel structures that are multiplications over dimensions and additions of base kernels inside dimensions. For example, on two dimensions we might consider
\begin{align*}
(\mathrm{SE}_{1}+\mathrm{PER}_{1})\times \mathrm{SE}_{2}.
\end{align*}
As addition is commutative it should not make a difference for the values of the predicted parameters if we change the order of the addition for example to 
\begin{align*}
(\mathrm{PER}_{1} + \mathrm{SE}_{1})\times \mathrm{SE}_{2}.
\end{align*}
This property is captured in our architecture as the predicted parameters of the base symbols are equivariant to such a permutation, thus only the index changes after the permutation (now  $\mathrm{SE}_{1}$ is the second entry in the addition) but not the value of the predicted parameter ($\hat{\theta}_{\mathrm{SE}_{1}}$ is unaffected from the permutation). 

A similar equivariance is captured in case the dimension index is reshuffled. A permutation of the input dimension effects both the dataset, that now has reshuffled dimensions indices, as well as the expression, that now also has reindexed dimensions. For example, swapping the two dimensions in our previous example leads to
 \begin{align*}
 \mathrm{SE}_{1} \times (\mathrm{SE}_{2}+\mathrm{PER}_{2}).
\end{align*}
Such a permutation should not change the previously predicted parameter value  $\hat{\theta}_{\mathrm{SE}_{2}}$ of the previous symbol $\mathrm{SE}_{2}$ that is now, after reindexing of the dimension, the predicted parameter $\hat{\theta}_{\mathrm{SE}_{1}}$ of symbol $\mathrm{SE}_{1}$. Here, our architecture is again equivariant meaning  that a permutation like that only changes the index of the symbol but not the value of the predicted parameter. We show these properties formally in Theorem \ref{main_theorem}.

Next, we give the proofs for the claimed invariances of our amortization network. First, we state the formal definition of an invariant and an equivariant function $f$ and prove some basic properties. 
\newtheorem{definition}{Definition}
\newtheorem{lemma}{Lemma}
\newtheorem{theorem}{Theorem}
\begin{definition}[Permutation Invariance] Let $\mathcal{G}$ be the permutation group and $f:\mathbb{X}^{N}\to\mathbb{Y}$ a function with $N\in\mathbb{N}$. The function $f$ is called permutation invariant iff
\begin{align*}
f(\mathbf{x})=f(x_{1},x_{2},\dots,x_{N})=f(x_{\pi(1)},x_{\pi(2)},\dots,x_{\pi(N)})=f(\pi \cdot \mathbf{x}),~~~ \forall \pi \in \mathcal{G}, \mathbf{x}\in\mathbb{X}^{N}.
\end{align*}
\end{definition}
\begin{definition}[Permutation Equivariant] Let $\mathcal{G}$ be the permutation group and $f:\mathbb{X}^{N}\to\mathbb{Y}^{N}$ a function with $N\in\mathbb{N}$. The function $f$ is called permutation equivariant iff
	\begin{align*}
	\pi \cdot f(\mathbf{x})=f(\pi \cdot \mathbf{x}),~~~ \forall \pi \in \mathcal{G}, \mathbf{x}\in\mathbb{X}^{N}.
	\end{align*}
\end{definition}
We first show that a function $f$ that maps each of its sequence elements $x_{i}\in\mathbb{X}$ element-wise via a function $g$ is permutation equivariant.
\begin{lemma}Let $f:\mathbb{X}^{N}\to\mathbb{Y}^{N}$ be a function with $N\in\mathbb{N}$ and let $f(x_{1},\dots,x_{N})=[g(x_{1}),\dots,g(x_{N})] \in \mathbb{Y}^{N} $ for some $g:\mathbb{X}\to\mathbb{Y}$. Then $f$ is permutation equivariant.
\end{lemma}
\textit{Proof:} Let $N\in \mathbb{N}$ and $\mathbf{x}=[x_{1},\dots,x_{N}] \in\mathbb{X}^{N}$ and let $\pi\in\mathcal{G}$. Let $y_{j}:=g(x_{j})$. Then
\begin{align*}
f(\pi \cdot \mathbf{x})&=f(x_{\pi(1)},x_{\pi(2)},\dots,x_{\pi(N)})=[g(x_{\pi(1)}),g(x_{\pi(2)}),\dots,g(x_{\pi(N)})]\\
&=[y_{\pi(1)},y_{\pi(2)},\dots,y_{\pi(N)}]=\pi \cdot [y_{1},\dots,y_{N}]=\pi \cdot f(\mathbf{x}).
\end{align*}
$\blacksquare$\\\\
Next, we show that chaining a permutation equivariant function with a permutation invariant function leads to a permutation invariant function.
\begin{lemma}Let $f:\mathbb{X}^{N}\to\mathbb{Y}^{N}$ be a permutation equivariant function with $N\in\mathbb{N}$ and let $g:\mathbb{Y}^{N}\to\mathbb{V}$ be permutation invariant. Then $h: \mathbb{X}^{N} \to \mathbb{V}$ with $h=g \circ f$ is  permutation invariant.
\end{lemma}
\textit{Proof:} Let $\mathbf{x}\in\mathbb{X}^{N}$ and $\pi \in\mathcal{G}$. Then
\begin{align*}
h(\pi \cdot \mathbf{x})=g(f(\pi \cdot \mathbf{x}))=g(\pi \cdot f(\mathbf{x}))=g(f(\mathbf{x}))=h(\mathbf{x}).
\end{align*}
$\blacksquare$\\\\
Furthermore, the chaining of two permutation equivariant functions is permutation equivariant.
\begin{lemma}Let $f:\mathbb{X}^{N}\to\mathbb{Y}^{N}$ be a permutation equivariant function with $N\in\mathbb{N}$ and let $g:\mathbb{Y}^{N}\to\mathbb{V}^{N}$ be permutation equivariant. Then $h: \mathbb{X}^{N} \to \mathbb{V}^{N}$ with $h=g \circ f$ is  permutation equivariant.
\end{lemma}
\textit{Proof:} Let $\mathbf{x}\in\mathbb{X}^{N}$ and $\pi \in\mathcal{G}$. Then
\begin{align*}
h(\pi \cdot \mathbf{x})=g(f(\pi \cdot \mathbf{x}))=g(\pi \cdot f(\mathbf{x}))=\pi \cdot g(f(\mathbf{x}))= \pi \cdot h(\mathbf{x}).
\end{align*}
$\blacksquare$\\\\
Furthermore, the chaining of a permutation invariant function with any function is permutation invariant.
\begin{lemma}Let $f:\mathbb{X}^{N}\to\mathbb{Y}$ be a permutation invariant function with $N\in\mathbb{N}$ and let $g:\mathbb{Y}\to\mathbb{V}$. Then $h: \mathbb{X}^{N} \to \mathbb{V}$ with $h=g \circ f$ is  permutation invariant.
\end{lemma}
\textit{Proof:} Let $\mathbf{x}\in\mathbb{X}^{N}$ and $\pi \in\mathcal{G}$. Then
\begin{align*}
h(\pi \cdot \mathbf{x})=g(f(\pi \cdot \mathbf{x}))=g(f(\mathbf{x}))=h(\mathbf{x}).
\end{align*}
$\blacksquare$\\\\
Furthermore, we make use of the following mappings that are used in our architecture:
\begin{enumerate}
	\item \textbf{Transformer-Encoder without positional encoding}: We use the encoder part of the transformer architecture without positional-encoding and without dropout \citep{AttentionIsAllYouNeed,SetTransformer} and denote the mapping as $\mathbf{Transformer}:\mathbb{X}^{n}\to\mathbb{X}^{n}$ with $\mathbb{X}\subset \mathbb{R}^{h}$. This block uses a stack of multi-head-self-attention layer \citep{AttentionIsAllYouNeed} denoted as mapping $\mathbf{MHSA}:\mathbb{X}^{n}\to\mathbb{X}^{n}$. Both, the $\textbf{MHSA}$ (without dropout) and $\mathbf{Transformer}$ (without positional encoding and dropout) are permutation-equivariant as shown in \cite{SetTransformer} Property 1 and Section 3.4.
	\item \textbf{Mean aggregation:} The mean aggregation $\mathbf{MeanAGG}:\mathbb{X}^{N}\to \mathbb{X}$ with $\mathbb{X}\subset \mathbb{R}^{h}$ is permutation invariant.
\end{enumerate}
\subsection{Main Invariances and Equivariances}
Next, we will prove the mentioned invariances for the several parts of the amortization network. We denote the dataset with $\mathcal{D}=\{(x_{j},y_{j})\in\mathbb{R}^{d+1}|j=1,\dots,n\}=\{(x_{j}^{(1)},\dots,x_{j}^{(d)},y_{j})\in\mathbb{R}^{d+1}|j=1,\dots,n\}$. A permutation $\pi$ of the dataset elements is then a shuffling of $[(x_{1},y_{1}),\dots,(x_{n},y_{n})]$ to $[(x_{\pi(1)},y_{\pi(1)}),\dots,(x_{\pi(n)},y_{\pi(n)})]$. We denote the reindexed dataset with $\mathcal{D}_{\pi}$. A permutation $\hat{\pi}$ of the dimensions is a reshuffling of $\mathcal{D}=\{(x_{j}^{(1)},\dots,x_{j}^{(d)},y_{j})\in\mathbb{R}^{d+1}|j=1,\dots,n\}$ to $\{(x_{j}^{(\hat{\pi}(1))},\dots,x_{j}^{(\hat{\pi}(d))},y_{j})\in\mathbb{R}^{d+1}|j=1,\dots,n\}$. We denote the resulting dataset with $\mathcal{D}_{\hat{\pi}}$.
\begin{theorem}
	\label{dataset_enc_theorem}
	Let $\mathcal{D}$ be a dataset with $n\in\mathbb{N}$ elements and $d\in\mathbb{N}$ input-dimensions. The dataset-encoder $g_{D}$ as a mapping from the dataset $\mathcal{D}$ to embedding vectors $[\mathbf{h}_{1},\dots,\mathbf{h}_{d}]$ (described through the steps 1.-9.) is permutation-invariant to a shuffling of the dataset elements (in the sense that $\mathbf{h}_{i}$ is invariant for all $i=1,\dots,d$) and permutation-equivariant to a shuffling of the dimensions.
\end{theorem}
\textit{Proof:} 
We go along the computation steps 1.-9. and keep track of the invariances/equivariances of the respective quantities. We note that applying multiple equivariant transformations in a row, keeps the equivariance according to Lemma 3. For each quantity we add the respective computation step index as right-most lower index to each quantity. We also do that with the weights of the respective learnable function.
\begin{enumerate}
	\item  Here, the dataset $\mathcal{D}$ is transformed into dimension-wise sequences $\mathcal{H}^{(i)}=[(x_{j}^{(i)},y_{j})]_{j=1}^{n}$ where we denote $h_{j,1}^{(i)}:=(x_{j}^{(i)},y_{j})$.

	\textit{Dataset shuffling:} When we shuffle the dataset elements via permutation $\pi$ this translates to shuffeled sequences $[(x_{\pi(j)}^{(i)},y_{\pi(j)})]_{j=1}^{n}$ applied to all dimensions $i=1,\dots,d$. Thus, the sequences $[h_{j,1}^{(i)}]_{j=1}^{n}$ are equivariant to a dataset shuffling.
	
	\textit{Dimension shuffling:} When we shuffle the dimension via a permutation $\hat{\pi}$ this translates to the shuffling of the sequence of sequences $[\mathcal{H}^{(1)},\dots,\mathcal{H}^{(d)}]$ to $[\mathcal{H}^{(\hat{\pi}(1))},\dots,\mathcal{H}^{(\hat{\pi}(d))}]$ by definition of $\mathcal{H}^{(i)}$. The sequence of sequences itself $[\mathcal{H}^{(1)}_{1},\dots,\mathcal{H}^{(d)}_{1}]$ is thus equivariant, to a shuffling of the input dimensions.
	
	\item Each element $h_{j,1}^{(i)}$ in $\mathcal{H}^{(i)}_{1}$ is mapped element-wise via the same linear layer $\mathbf{Linear}_{W_{2}}: \mathbb{R}^{2}\to \mathbb{R}^{h}$ with weights $W_2$ to $h_{j,2}^{(i)}$:
	\begin{align*}
		h_{j,2}^{(i)}=\mathbf{Linear}_{W_{2}}\bigg(h_{j,1}^{(i)}\bigg).
	\end{align*}
	 \textit{Dataset shuffling:} From Lemma 1 with $g=\mathbf{Linear}_{W_{2}}$ it follows that the mapped sequence $\mathcal{H}^{(i)}_{2}=[h_{1,2}^{(i)},\dots,h_{n,2}^{(i)}]$ is equivariant to a dataset shuffling for all $i=1,\dots,d$.
	 
	 \textit{Dimension shuffling:} As the same mapping is applied for all dimensions $i=1,\dots,d$ via
	 	\begin{align*}
	 	\mathcal{H}^{(i)}_{2}=\mathbf{ElemLinear}_{W_{2}}(\mathcal{H}^{(i)}_{1}):=[\mathbf{Linear}_{W_{2}}(h_{1,1}^{(i)}),\dots,\mathbf{Linear}_{W_{2}}(h_{n,1}^{(i)})],
	 	\end{align*}
	  the sequence of transformed sequences $[\mathcal{H}^{(1)}_{2},\dots,\mathcal{H}^{(d)}_{2}]=[\mathbf{ElemLinear}_{W_{2}}(\mathcal{H}^{(1)}_{1}),\dots,\mathbf{ElemLinear}_{W_{2}}(\mathcal{H}^{(d)}_{1})]$ is equivariant to dimension shuffling with Lemma 1 and $g=\mathbf{ElemLinear}_{W_{2}}$.
	\item Here, the $\textbf{Transformer}_{W_{3}}$ block with weights $W_{3}$ is applied and results in  the transformed sequence $\mathcal{H}^{(i)}_{3}:=[h_{1,3}^{(i)},\dots,h_{n,3}^{(i)}] = \textbf{Transformer}_{W_{3}}([h_{1,2}^{(i)},\dots,h_{n,2}^{(i)}])$.
	
	 \textit{Dataset shuffling:} As the $\textbf{Transformer}_{W_{3}}$ block is permutation equivariant, the transformed sequence $\mathcal{H}^{(i)}_{3}=[h_{1,3}^{(i)},\dots,h_{n,3}^{(i)}] = \textbf{Transformer}_{W_{3}}([h_{1,2}^{(i)},\dots,h_{n,2}^{(i)}])$ is equivariant to dataset shuffling for all $i=1,\dots,d$. 
	 
	 \textit{Dimension shuffling:} As we apply the same (weight-shared) transformer block $\textbf{Transformer}_{W_{3}}$ to all dimensions $i=1,\dots,d$, and applying Lemma 1 with $g=\textbf{Transformer}_{W_{3}}$, the transformed sequence of sequences
	\begin{align*}
	[\mathcal{H}^{(1)}_{3},\dots,\mathcal{H}^{(d)}_{3}]=[\textbf{Transformer}_{W_{3}}(\mathcal{H}^{(1)}_{2}),\dots,\textbf{Transformer}_{W_{3}}(\mathcal{H}^{(d)}_{2})]
	\end{align*}
	is equivariant to input dimension shuffling.
	\item Here, we first construct the combined (over $i=1,\dots,d$) sequences $\mathcal{W}_{j,4}:=[h_{j,3}^{(1)},\dots,h_{j,3}^{(d)}]$ and apply mean aggregation on them via
		\begin{align*}
		h_{j,4}:=\textbf{MeanAGG}(\mathcal{W}_{j,4})=\textbf{MeanAGG}([h_{j,3}^{(1)},\dots,h_{j,3}^{(d)}])
		\end{align*}
	 to form the sequence $[h_{1,4},\dots,h_{n,4}]$.
	 
	\textit{Dataset shuffling:} As $[h_{1,3}^{(i)},\dots,h_{n,3}^{(i)}]$ is equivariant to dataset shuffling for all $i=1,\dots,d$, also the sequence $[\mathcal{W}_{1,4},\dots,\mathcal{W}_{n,4}]$  of the combined (over $i=1,\dots,d$) elements $\mathcal{W}_{j,4}:=[h_{j,3}^{(1)},\dots,h_{j,3}^{(d)}]$ is equivariant to dataset shuffling. As we apply the same function to each $\mathcal{W}_{j,4}$, via $h_{j,4}:=\textbf{MeanAGG}(\mathcal{W}_{j,4})$, also the sequence $[h_{1,4},\dots,h_{n,4}]$ is equivariant to dataset shuffling via Lemma 1.
	
	 \textit{Dimension shuffling:} As $[\mathcal{H}^{(1)}_{3},\dots,\mathcal{H}^{(d)}_{3}]$ is equivariant to dimension shuffling each sequence $[h_{j,3}^{(1)},\dots,h_{j,3}^{(d)}]$ is equivariant to dimension shuffling for all $j=1,\dots,n$. Applying the mean aggregation $h_{j,4}=\textbf{MeanAGG}([h_{j,3}^{(1)},\dots,h_{j,3}^{(d)}])$ onto $[h_{j,3}^{(1)},\dots,h_{j,3}^{(d)}]$ renders via Lemma 2 the quantity $h_{j,4}$ invariant to dimension shuffling and thus also the complete sequence $[h_{1,4},\dots,h_{n,4}]$.
	 
	\item Here, we apply the transformer block $\textbf{Transformer}_{W_{5}}$ with weights $W_{5}$ on $[h_{1,4},\dots,h_{n,4}]$ to form the sequence 
	\begin{align*}
	[h_{1,5},\dots,h_{n,5}] = \mathbf{Transformer}_{W_{5}}([h_{1,4},\dots,h_{n,4}]).
	\end{align*}
	
	 \textit{Dataset shuffling:} As $[h_{1,4},\dots,h_{n,4}]$ is equivariant to dataset shuffling and the transformer block $\textbf{Transformer}_{W_{5}}$ with weights $W_{5}$ keeps equivariance also the transformed sequence $[h_{1,5},\dots,h_{n,5}] $ is equivariant to dataset shuffling.
	 
	 \textit{Dimension shuffling:} The previous sequence $[h_{1,4},\dots,h_{n,4}]$ is invariant to dimension shuffling, and the invariance of  $[h_{1,5},\dots,h_{n,5}]$ to dimension shuffling is preserved via Lemma 4.
	\item Here, we concatenate for each $i=1,\dots,d$ the sequences $\mathcal{H}^{(i)}_{3}=[h_{1,3}^{(i)},\dots,h_{n,3}^{(i)}]$ from step 3 with $[h_{1,5},\dots,h_{n,5}]$ via 
	\begin{align*}
	h_{j,6}^{(i)}:=\mathbf{Concat}(h_{j,5},h_{j,3}^{(i)})
	\end{align*}
	to form the combined sequences $\mathcal{H}^{(i)}_{6}:=[h_{1,6}^{(i)},\dots,h_{n,6}^{(i)}]$ with  $i=1,\dots,d$.
	
	 \textit{Dataset shuffling:} As the sequences $\mathcal{H}^{(i)}_{3}=[h_{1,3}^{(i)},\dots,h_{n,3}^{(i)}]$ from step 3. are equivariant to a shuffling of the dataset elements for all $i=1,\dots,d$ as well as the sequence $[h_{1,5},\dots,h_{n,5}]$ from step 5 also the concatenation of the elements of the two sequences is permutation equivariant. Thus, the transformed sequence $\mathcal{H}^{(i)}_{6}:=[h_{1,6}^{(i)},\dots,h_{n,6}^{(i)}]$ is permutation equivariant to dataset shuffling for all $i=1,\dots,d$. 
	 
	 \textit{Dimension shuffling:} The sequence of transformed sequences $[\mathcal{H}^{(1)}_{6},\dots,\mathcal{H}^{(d)}_{6}]$ is permutation equivariant to a shuffling of the dimensions as we applied the same function to all sequences, via 
	 \begin{align*}
	 \mathcal{H}^{(i)}_{6}= \mathbf{ElemConcat}([h_{j,5}]_{j=1}^{n},\mathcal{H}^{(i)}_{3})=[\mathbf{Concat}(h_{1,5},h_{1,3}^{(i)}),\dots,\mathbf{Concat}(h_{n,5},h_{n,3}^{(i)})],
	 \end{align*}
	  where $[h_{j,5}]_{j=1}^{n}$ is invariant to the shuffling of the dimensions (see step 5).
	\item Here, we apply the $\textbf{Transformer}_{W_{7}}$ block with weights $W_{7}$ to get the transformed sequence 
	\begin{align*}
	[h_{1,7}^{(i)},\dots,h_{n,7}^{(i)}] = \textbf{Transformer}_{W_{7}}([h_{1,6}^{(i)},\dots,h_{n,6}^{(i)}]).
	\end{align*}
	
	 \textit{Dataset shuffling:} The $\textbf{Transformer}_{W_{7}}$ block with weights $W_{7}$ is equivariant, therefore is the transformed sequence $[h_{1,7}^{(i)},\dots,h_{n,7}^{(i)}]$ equivariant to dataset shuffling for all $i=1,\dots,d$.
	 
	 \textit{Dimension shuffling:} As we use the same transformer for all dimensions $i=1,\dots,d$, the transformed sequence of sequences \begin{align*}
	[\mathcal{H}^{(1)}_{7},\dots,\mathcal{H}^{(d)}_{7}]=[\textbf{Transformer}_{W_{7}}(\mathcal{H}^{(1)}_{6}),\dots,\textbf{Transformer}_{W_{7}}(\mathcal{H}^{(d)}_{6})]
	\end{align*} 
	keeps its equivariance to the shuffling of the input dimensions.
	\item Here, we apply the mean aggregation
	\begin{align*}
	\mathbf{h}_{i,8}=\textbf{MeanAGG}([h_{1,7}^{(i)},\dots,h_{n,7}^{(i)}])
	\end{align*}
	 on each sequence $\mathcal{H}^{(i)}_{7}=[h_{1,7}^{(i)},\dots,h_{n,7}^{(i)}]$ to form the compressed sequence $[\mathbf{h}_{1,8},\dots,\mathbf{h}_{d,8}]$
	
	\textit{Dataset shuffling:} As the mean aggregation is invariant to shuffling and the sequences $\mathcal{H}^{(i)}_{7}=[h_{1,7}^{(i)},\dots,h_{n,7}^{(i)}]$ are equivariant to a shuffling of the dataset elements, with Lemma 2, it follows that the vector $\mathbf{h}_{i,8}=\textbf{MeanAGG}([h_{1,7}^{(i)},\dots,h_{n,7}^{(i)}])$ is invariant to the shuffling of the dataset elements and thus also the complete sequence $[\mathbf{h}_{1,8},\dots,\mathbf{h}_{d,8}]$. 
	
	\textit{Dimension shuffling:} As we apply the same function (mean aggregation) to all sequences $\mathcal{H}^{(i)}_{7},i=1,\dots,d$ and the sequence of sequences $[\mathcal{H}^{(1)}_{7},\dots,\mathcal{H}^{(d)}_{7}]$ is permutation equivariant to dimension shuffling, also the output embeddings sequence $[\mathbf{h}_{1,8},\dots,\mathbf{h}_{d,8}]$ is permutation equivariant to dimension shuffling.
	\item In the last step we apply a $\mathbf{Transformer}_{W_{9}}$ with weights $W_{9}$ to $[\mathbf{h}_{1,8},\dots,\mathbf{h}_{d,8}]$ to get the sequence  \begin{align*}[\mathbf{h}_{1,9},\dots,\mathbf{h}_{d,9}]= \mathbf{Transformer}_{W_{9}}([\mathbf{h}_{1,8},\dots,\mathbf{h}_{d,8}]).\end{align*}
	
	\textit{Dataset shuffling:} As $[\mathbf{h}_{1,8},\dots,\mathbf{h}_{d,8}]$ was invariant to to dataset shuffling before the transformation, it is still after the transformation, via Lemma 4.
	
	\textit{Dimension shuffling:} As the transformer block $\mathbf{Transformer}_{W_{9}}$ with weights $W_{9}$ is equivariant and the input sequence  $[\mathbf{h}_{1,8},\dots,\mathbf{h}_{d,8}]$ is equivariant, also the sequence  $[\mathbf{h}_{1,9},\dots,\mathbf{h}_{d,9}]$ is equivariant to a shuffling of the input dimensions. $\blacksquare$
\end{enumerate}
Next, we will prove invariances of the kernel-encoder-decoder block $g_k(\mathbf{h}_{\mathcal{D}},\mathcal{V}_{\mathcal{S}})$ that maps a sequence of sequences $\mathcal{V}_{\mathcal{S}}=[\mathcal{V}_{1},\dots,\mathcal{V}_{d}]$ of base kernel representations, with $\mathcal{V}_{i}=[v_{1}^{(i)},\dots,v_{N_{i}}^{(i)}]$ and the sequence of dataset embeddings $[\mathbf{h}_{1},\dots,\mathbf{h}_{d}]$ to a transformed sequence of sequences $[\tilde{\mathcal{V}}_{1},\dots,\tilde{\mathcal{V}}_{d}]$ of learned base kernel representations. In particular, when we refer to a permutation $\hat{\pi}$ of the input dimensions of the datset we implicity assume that the input sequence $[\mathcal{V}_{1},\dots,\mathcal{V}_{d}]$ is shuffled in the same way, thus to $[\mathcal{V}_{\hat{\pi}(1)},\dots,\mathcal{V}_{\hat{\pi}(d)}]$. Furthermore, as one sequence $\mathcal{V}_{i}=[v_{1}^{(i)},\dots,v_{N_{i}}^{(i)}]$ describes an addition of kernels we also consider the permutation equivariance of these subsequences in the sense that for any dimension $i\in\{1,\dots,n\}$ the output sequence, denoted by $\tilde{\mathcal{V}}_{i}=[\tilde{v}_{1}^{(i)},\dots,\tilde{v}_{N_{i}}^{(i)}]$ is permutation equivariant to a permutation $\tilde{\pi}$ of the input sequence $\mathcal{V}_{i}=[v_{1}^{(i)},\dots,v_{N_{i}}^{(i)}]$. We call it permutation of the base-symbols within dimension $i$.

 First, we will show that the \textbf{Kernel-Encoder-Block}, as a function from a sequence $[v_{1},\dots,v_{N}]$ and a context $c$ to a transformed sequence $[v_{1},\dots,v_{N}]$, is permutation equivariant given the context $c$.
\begin{theorem}
	\label{kernel_enc_theorem}
	The \textbf{Kernel-Encoder-Block} $g$ as described in Figure 1 b) in the main paper is permutation equivariant as a mapping from a sequence $[v_{1},\dots,v_{N}]$ with $v_{j}\in\mathbb{R}^{h}$ to a sequence $[v_{1},\dots,v_{N}]$ given a context vector $c\in\mathbb{R}^{l}$, meaning for a permutation $\pi \in \mathcal{G}$ and a context vector  $c\in\mathbb{R}^{l}$:
	\begin{align*}
		g(\pi \cdot [v_{1},\dots,v_{N}],c)= \pi \cdot g([v_{1},\dots,v_{N}],c).
	\end{align*}
\end{theorem}
\textit{Proof:} The \textbf{Kernel-Encoder-Block} first maps the sequence through a multi-head-self-attention $\textbf{MHSA}$ layer with output sequence $[\tilde{v}_{1},\dots,\tilde{v}_{N}]$. After that it adds the input sequence $[v_{1},\dots,v_{N}]$ element-wise to the output sequence and applies a layer normalization step element-wise. We denote the output of that operation as $[\bar{v}_{1},\dots,\bar{v}_{N}]$. This sequence is concatenated with the context vector to  $[(\bar{v}_{1},c),\dots,(\bar{v}_{N},c)]$. This is given element-wise to a shared \textbf{MLP} layer, where we denote the output with $[\hat{v}_{1},\dots,\hat{v}_{N}]$. This is followed by an addition of $[\bar{v}_{1},\dots,\bar{v}_{N}]$ to $[\hat{v}_{1},\dots,\hat{v}_{N}]$ and a layer normalization step that is applied element-wise. \\\\
As  $\textbf{MHSA}$ is permutation equivariant the sequence $[\tilde{v}_{1},\dots,\tilde{v}_{N}]$ is permutation equivariant and thus also the addition to the original sequence $[v_{1},\dots,v_{N}]$. As the layer normalizaton is applied element-wise, via Lemma 1, the sequence $[\bar{v}_{1},\dots,\bar{v}_{N}]$ is permutation equivariant. As we concatenate equivariant elements $\bar{v}_{j}$ with a fixed quantity $c$, the sequence  $[(\bar{v}_{1},c),\dots,(\bar{v}_{N},c)]$ is permutation equivariant to a shuffling of $[v_{1},\dots,v_{N}]$. The $\textbf{MLP}_{W}$ with weights $W$ is applied element-wise:
\begin{align*}
[\hat{v}_{1},\dots,\hat{v}_{N}]=[\textbf{MLP}_{W}(\bar{v}_{1},c),\dots,\textbf{MLP}_{W}(\bar{v}_{N},c)]
\end{align*}
and via Lemma 1 it holds that the output sequence is permutation-equivariant. As $[\bar{v}_{1},\dots,\bar{v}_{N}]$ is permutation equivariant as well as $[\hat{v}_{1},\dots,\hat{v}_{N}]$ their addition is permutation equivariant. As the final layer normalization is applied element-wise, via Lemma 1, the final output is permutation equivariant.$\blacksquare$\\\\
Next, we prove the mentioned invariances for the kernel-encoder-decoder $g_k$.
\begin{theorem}
	\label{kernel_enc_dec_theorem}
	Let $[\mathbf{h}_{1},\dots,\mathbf{h}_{d}]$ be the output of $g_{D}$ for the dataset $\mathcal{D}$ and let $\mathcal{V}_{\mathcal{S}}=[\mathcal{V}_{1},\dots,\mathcal{V}_{d}]$ be the sequence of sequences of input base-symbol encodings. The output sequence of sequences $[\tilde{\mathcal{V}}_{1},\dots,\tilde{\mathcal{V}}_{d}]$ of the kernel-encoder-decoder $g_k(\mathbf{h}_{\mathcal{D}},\mathcal{V}_{\mathcal{S}})$ is permutation equivariant to a shuffling of the dimensions. Furthermore, for any dimension $i\in\{1,\dots,d\}$ the single output sequence $\tilde{\mathcal{V}}_{i}=[\tilde{v}_{1}^{(i)},\dots,\tilde{v}_{N_{i}}^{(i)}]$ is equivariant to a permutation $\pi$ of the respective input sequence $\mathcal{V}_{i}=[v_{1}^{(i)},\dots,v_{N_{i}}^{(i)}]$.
\end{theorem}
\textit{Proofs:} 
First, we show that a stacked version of the \textbf{Kernel-Encoder-Block} is also permutation equivariant given a context (just as a single \textbf{Kernel-Encoder-Block} according to Theorem \ref{kernel_enc_theorem}). To show that, we denote the \textbf{Kernel-Encoder-Block} as the mapping $\mathbf{KernelEncoder}_{W}: \mathbb{X}^{N}\times \mathbb{R}^{l}\to \mathbb{X}^{N}$ with weights $W$ and a stacked version with 
\begin{align*}
\mathbf{KernelEncoderStacked}(\mathcal{V},c):=\mathbf{KernelEncoder}_{\tilde{W}_{1}}(\mathbf{KernelEncoder}_{\tilde{W}_{2}}(\mathcal{V},c),c)
\end{align*}
 here w.l.o.g with two layers and weights $\tilde{W}_{1},\tilde{W}_{2}$. It is permutation equivariant given a context $c\in\mathbb{R}^{h}$ with $h\in\mathbb{N}$ since
\begin{align*}
\pi\mathbf{KernelEncoderStacked}(\mathcal{V},c)&=\pi\mathbf{KernelEncoder}_{\tilde{W}_{1}}(\mathbf{KernelEncoder}_{\tilde{W}_{2}}(\mathcal{V},c),c)\\&=\mathbf{KernelEncoder}_{\tilde{W}_{1}}(\pi\mathbf{KernelEncoder}_{\tilde{W}_{2}}(\mathcal{V},c),c)\\&=\mathbf{KernelEncoder}_{\tilde{W}_{1}}(\mathbf{KernelEncoder}_{\tilde{W}_{2}}(\pi\mathcal{V},c),c).
\end{align*}
We denote a stack of \textbf{Kernel-Encoder-Block} layers with weights $\tilde{W}_{1},\dots,\tilde{W}_{k}$ with $\mathbf{KernelEncoderStacked}_{W}$ where we summarize all weights to $W$.

Now, let $\mathcal{D}$ be a dataset with $n\in \mathbb{N}$ datapoints and $d\in \mathbb{N}$ input-dimensions. Let $[\mathbf{h}_{1},\dots,\mathbf{h}_{d}]=g_{D}(\mathcal{D})$ and let $\mathcal{V}_{\mathcal{S}}=[\mathcal{V}_{1},\dots,\mathcal{V}_{d}]$ be the sequence of sequences of input base-symbol encodings with $\mathcal{V}_{i}=[v_{1}^{(i)},\dots,v_{N_{i}}^{(i)}]$. Furthermore, let $l\in\{1,\dots,d\}$ be the dimension on which we apply the permutation  $\tilde{\pi}$ of the base symbol sequence $\mathcal{V}_{l}=[v_{1}^{(l)},\dots,v_{N_{l}}^{(l)}]$.

 Similar as in Theorem \ref{dataset_enc_theorem} we now go along the computation steps 1.-4. of the kernel-encoder-decoder and consider the invariance/equivariances of the respective quantities. We add the respective computation step index as right-most lower index to each quantity and weight of the respective learnable function:
\begin{enumerate}
	\item Here, each sequence $\mathcal{V}_{i}=[v_{1}^{(i)},\dots,v_{N_{i}}^{(i)}], i=1,\dots,d$ is given to a $\mathbf{KernelEncoderStacked}_{W_{1}}$ block with weights $W_{1}$ together with context $\mathbf{h}_{i}$:
	\begin{align*}
	[v_{1,1}^{(i)},\dots,v_{N_{i},1}^{(i)}]=\mathbf{KernelEncoderStacked}_{W}([v_{1}^{(i)},\dots,v_{N_{i}}^{(i)}],\mathbf{h}_{i})
	\end{align*}
	 \textit{Shuffling of base symbols in dimension $l$:} As $\mathbf{KernelEncoderStacked}_{W_{1}}$ is equivariant given a context (and we set the context to the fixed quantity $\mathbf{h}_{l}$), the transformed sequence  $\mathcal{V}_{l,1}=[v_{1,1}^{(l)},\dots,v_{N_{l},1}^{(l)}]$ is permutation equivariant to a shuffling of the base-symbols in dimension $l$. 
	 
	 \textit{Shuffling of dimension:} The sequence $[\mathbf{h}_{1},\dots,\mathbf{h}_{d}]$ is permutation equivariant to a shuffling of the input dimensions according to Theorem \ref{dataset_enc_theorem} and the sequence $[\mathcal{V}_{1},\dots,\mathcal{V}_{d}]$ by definition. Thus, the sequence of tuples $[(\mathcal{V}_{1},\mathbf{h}_{1}),\dots,(\mathcal{V}_{d},\mathbf{h}_{d})]$ is equivariant to a shuffling of the dimension and we apply the same $\mathbf{KernelEncoderStacked}_{W_{1}}$ block to each tuple with
	 \begin{align*}
	 [\mathcal{V}_{1,1},\dots,\mathcal{V}_{d,1}]=[\mathbf{KernelEncoderStacked}_{W_{1}}(\mathcal{V}_{1},\mathbf{h}_{1}),\dots,\mathbf{KernelEncoderStacked}_{W_{1}}(\mathcal{V}_{d},\mathbf{h}_{d})].
	 \end{align*}
	 Thus, the transformed sequence of sequences $[\mathcal{V}_{1,1},\dots,\mathcal{V}_{d,1}]$ is equivariant to dimension shuffling, according to Lemma 1 with $g=\mathbf{KernelEncoderStacked}_{W_{1}}$.
	\item Here, we apply a mean aggregation on the sequences $\mathcal{V}_{i,1}=[v_{1,1}^{(i)},\dots,v_{N_{i},1}^{(i)}],i=1,\dots,d$ via 
	\begin{align*}\mathbf{v}_{i,2}=\textbf{MeanAGG}([v_{1,1}^{(i)},\dots,v_{N_{i},1}^{(i)}])\end{align*}
	to construct the sequence $[\mathbf{v}_{1,2},\dots,\mathbf{v}_{d,2}]$.
	
	 \textit{Shuffling of base symbols in dimension $l$:} As the sequence $[v_{1,1}^{(l)},\dots,v_{N_{l},1}^{(l)}]$ is equivariant to a shuffling of the base-symbols in dimension $l$ and the mean aggregation is permutation invariant, the kernel embedding in dimension $l$, given as $\mathbf{v}_{l,2}=\textbf{MeanAGG}([v_{1,1}^{(l)},\dots,v_{N_{l},1}^{(l)}])$, is invariant to a shuffling of the base-symbols in dimension $l$ via Lemma 2. For any other dimension $i\neq l$ the quantity $\mathbf{v}_{i,2}$ is also invariant to a shuffling of base symbols in dimension $l$, as the permuted sequence $\mathcal{V}_{l}$ is not an input to that quantity. Thus, the complete sequence  $[\mathbf{v}_{1,2},\dots,\mathbf{v}_{d,2}]$ is invariant to a shuffling of base symbols in dimension $l$. 
	 
	 \textit{Shuffling of dimension:} Furthermore, the sequence of embeddings $[\mathbf{v}_{1,2},\dots,\mathbf{v}_{d,2}]$ is equivariant to dimension shuffling as we apply the same mapping onto the equivariant sequence $[\mathcal{V}_{1,1},\dots,\mathcal{V}_{d,1}]$ via
	\begin{align*}
	[\mathbf{v}_{1,2},\dots,\mathbf{v}_{d,2}]=[\textbf{MeanAGG}(\mathcal{V}_{1,1}),\dots,\textbf{MeanAGG}(\mathcal{V}_{d,1})]
	\end{align*}
	and applying Lemma 1 with $g=\textbf{MeanAGG}$.
	
	\item Here, we apply a $\mathbf{Transformer}_{W_{3}}$ with weights $W_{3}$ to get the transformed sequence  \begin{align*}[\mathbf{v}_{1,3},\dots,\mathbf{v}_{d,3}]= \mathbf{Transformer}_{W_{3}}([\mathbf{v}_{1,2},\dots,\mathbf{v}_{d,2}]).\end{align*}
	
	\textit{Shuffling of base symbols in dimension $l$:}
	As $[\mathbf{v}_{1,2},\dots,\mathbf{v}_{d,2}]$ is invariant to a shuffling of base symbols in dimension $l$, also the transformation $[\mathbf{v}_{1,3},\dots,\mathbf{v}_{d,3}]$ is, via Lemma 4.

	\textit{Shuffling of dimension:}
	 As the transformer block $\mathbf{Transformer}_{W_{3}}$ is equivariant, the equivariance to dimension shuffling is preserved for the transformed sequence  $[\mathbf{v}_{1,3},\dots,\mathbf{v}_{d,3}]$ 
	 
	\item In the last step, each sequence $\mathcal{V}_{i,1}=[v_{1,1}^{(i)},\dots,v_{N_{i},1}^{(i)}]$ from step 1 with $i=1,\dots,d$ is given to a $\mathbf{KernelEncoderStacked}_{W_4}$ block with weights $W_{4}$:
	\begin{align*}
	[v_{1,4}^{(i)},\dots,v_{N_{i},4}^{(i)}]=\mathbf{KernelEncoderStacked}_{W_4}([v_{1,1}^{(i)},\dots,v_{N_{i},1}^{(i)}],c_{i,4})
	\end{align*} together with the extended context $c_{i,4}$, where 
	\begin{align*}
	c_{i,4}=\mathbf{Concat}(\mathbf{h}_{i},\mathbf{v}_{i,3}).
	\end{align*}
	
	\textit{Shuffling of base symbols in dimension $l$:} As $\mathbf{v}_{l,3}$ is invariant to a base-symbol shuffling within dimension $l$ also the concatenated context $c_{l,4}=\mathbf{Concat}(\mathbf{h}_{l},\mathbf{v}_{l,3})$ is invariant to that shuffling. 
	Thus, independent of the shuffling of the base-symbols within the dimension $l$, the context vectors stays the same and we can apply Theorem \ref{kernel_enc_theorem}. Thus, via Theorem \ref{kernel_enc_theorem} the $\mathbf{KernelEncoderStacked}_{W_4}$ is equivariant given the context $c_{l,4}$, therefore, with the fact that $[v_{1,1}^{(l)},\dots,v_{N_{l},1}^{(l)}]$ is equivariant to a shuffling of the base symbols in dimension $l$ we conclude that also the transformed sequence $\mathcal{V}_{l,4}=[v_{1,4}^{(l)},\dots,v_{N_{l},4}^{(l)}]$ is permutation equivariant to a shuffling of the base-symbols in dimension $l$.

	\textit{Shuffling of dimension:} The sequence of input tuples $[(\mathcal{V}_{1,1},c_{1,4}),\dots,(\mathcal{V}_{d,1},c_{d,4})]$ is equivariant to a shuffling of the dimension as $[c_{1,4},\dots,c_{d,4}]$ and $[\mathcal{V}_{1,1},\dots,\mathcal{V}_{d,1}]$ are equivariant. Applying the same $\mathbf{KernelEncoderStacked}_{W_4}$ block to each tuple renders the transformed sequence \begin{align*}[\mathcal{V}_{1,4},\dots,\mathcal{V}_{d,4}]=[\mathbf{KernelEncoderStacked}_{W_4}(\mathcal{V}_{1,1},c_{1,4}),\dots,\mathbf{KernelEncoderStacked}_{W_4}(\mathcal{V}_{d,1},c_{d,4})] \end{align*} equivariant to dimension shuffling, according to Lemma 1 with $g=\mathbf{KernelEncoderStacked}_{W_4}$. $\blacksquare$
\end{enumerate}

\paragraph{Noise Variance Predictor:} We give a short detailed notation for the noise variance predictor. The noise variance predictor $g_{NV}$ gets as input the output of the dataset encoder $[\mathbf{h}_{i}]_{i=1}^{d}$ and the output of the kernel-encoder-decoder $\tilde{V}_{\mathcal{S}}=[\tilde{\mathcal{V}}_{1},\dots,\tilde{\mathcal{V}}_{d}]$ and maps to a single real value $\hat{\sigma}^{2}=g_{NV}([\mathbf{h}_{i}]_{i=1}^{d},\tilde{V}_{\mathcal{S}})$ that is the prediction for the noise variance $\sigma^{2}$. It is defined via:
\begin{align*}
g_{NV}([\mathbf{h}_{i}]_{i=1}^{d},\tilde{V}_{\mathcal{S}})=\textbf{MLP}_{W}\bigg(\textbf{Concat}\bigg(\textbf{MeanAgg}([\mathbf{h}_{i}]_{i=1}^{d}),\textbf{MeanAgg}(\tilde{V}_{\mathcal{S}})\bigg)\bigg)
\end{align*}
where $\textbf{MeanAgg}(\tilde{V}_{\mathcal{S}})$ is the mean over all $\tilde{v}_{j}^{(i)}$ in $\tilde{V}_{\mathcal{S}}$ and $W$ are the weights of the $\textbf{MLP}$ layer.

\paragraph{Kernel Parameter Predictor:} We give a short detailed notation for the kernel parameter predictor $g_{\Theta_{\mathcal{S}}}$. It gets as input the output of the kernel-encoder decoder $\tilde{V}_{\mathcal{S}}=[\tilde{\mathcal{V}}_{1},\dots,\tilde{\mathcal{V}}_{d}]$ and the sequence of sequences of base-symbols $\mathcal{B}_{\mathcal{S}}:=\bigg[[B_{1}^{(i)},\dots,B^{(i)}_{N_{i}}]|i=1,\dots,d\bigg]$ of $\mathcal{S}$ with $B_{j}^{(i)}\in \mathcal{B}$ and is defined via
\begin{align*}
g_{\Theta_{\mathcal{S}}}(\tilde{V}_{\mathcal{S}},\mathcal{B}_{\mathcal{S}}):=\bigg[[\textbf{MLP}_{B_{1}^{(i)}}(v_{1}^{(i)}),\dots,\textbf{MLP}_{B^{(i)}_{N_{i}}}(v_{N_{i}}^{(i)})]\bigg|i=1,\dots,d\bigg]
\end{align*}
where for each symbol $B\in\mathcal{B}$ in the corpus $\mathcal{B}$ a separate $\textbf{MLP}_{B}$ layer with weights $W_{B}$ is used.

The next theorem is our main theorem and combines the previous theorems to state the main invariances of the output quantities, namely the predicted kernel parameters and the predicted likelihood noise variance.
\begin{theorem}
	\label{main_theorem}
	Let $\mathcal{D}$ be a dataset with $d\in\mathbb{N}$ input dimensions and $n\in\mathbb{N}$ datapoints and $\mathcal{S}$ an expression as described in Section 2 of the paper. For the output of our amortization network $g(\mathcal{D}, \mathcal{S})$ given through the prediction of the kernel parameters $\theta_{B_{j}^{(i)}}\in\Theta_{B_{j}^{(i)}}$ for each base symbol in $\mathcal{S}$ with $i=1,\dots,d$ and $j=1,\dots,N_{i}$ and a prediction of the likelihood variance $\sigma^{2}$ the following properties hold:
	\begin{enumerate}
		\item The output is invariant to a shuffling of the dataset elements, meaning for any valid $\mathcal{S}$ and shuffeled dataset $\mathcal{D}_{\pi}$ it holds:
		\begin{align*}
		g(\mathcal{D},\mathcal{S})=g(\mathcal{D}_{\pi},\mathcal{S}).
		\end{align*}
		\item The output kernel parameters are equivariant to a dimension shuffling, meaning
		for a dimension shuffeled dataset $\mathcal{D}_{\hat{\pi}}$ and a dimension shuffeld expression $\mathcal{S}_{\hat{\pi}}$ with permuted sequence of sequence of base-kernels $\mathcal{B}_{\mathcal{S}_{\hat{\pi}}}$ it holds:
		\begin{align*}
		\hat{\theta}_{\hat{B}_{j}^{(i)}} = \theta_{B_{j}^{(\hat{\pi}(i))}} \in\Theta_{\hat{B}_{j}^{(i)}} = \Theta_{B_{j}^{(\hat{\pi}(i))}},
		\end{align*}
		where $\hat{\theta}_{\hat{B}_{j}^{(i)}} $ is the output of $g(\mathcal{D}_{\hat{\pi}},\mathcal{S}_{\hat{\pi}})$ for symbol $\hat{B}_{j}^{(i)}\in \mathcal{B}_{\mathcal{S}_{\hat{\pi}}}$ and $\theta_{B_{j}^{(i)}}$ is the output of $g(\mathcal{D},\mathcal{S})$ for symbol $B_{j}^{(i)}\in\mathcal{B}_{\mathcal{S}}$.
		\item For any $i\in\{1,\dots,d\}$ the output kernel parameters are equivariant to a permutation $\tilde{\pi}$ of the the base-symbols inside the subexpression $\mathcal{S}_{i}$ meaning for the shuffeled expression $\mathcal{S}_{\tilde{\pi}}$ with permuted sequence of sequence of base-kernels $\mathcal{B}_{\mathcal{S}_{\tilde{\pi}}}$ it holds for all $j=1,\dots,N_{i}$:
		\begin{align*}
\hat{\theta}_{\hat{B}_{j}^{(i)}} = \theta_{B_{\tilde{\pi}(j)}^{(i)}} \in\Theta_{\hat{B}_{j}^{(i)}} = \Theta_{B_{\tilde{\pi}(j)}^{(i)}},
\end{align*}
where $\hat{\theta}_{\hat{B}_{j}^{(i)}} $ is the output of $g(\mathcal{D},\mathcal{S}_{\tilde{\pi}})$ for symbol $\hat{B}_{j}^{(i)}\in \mathcal{B}_{\mathcal{S}_{\tilde{\pi}}}$ and $\theta_{B_{j}^{(i)}}$ is the output of $g(\mathcal{D},\mathcal{S})$ for symbol $B_{j}^{(i)}\in\mathcal{B}_{\mathcal{S}}$.
		\item The likelihood variance prediction is invariant to a permutation of the input dimension $\hat{\pi}$ and invariant to a permutation $\tilde{\pi}$ of the the base-symbols inside a subexpression $\mathcal{S}_{i}$ applied to any input dimension $i\in\{1,\dots,d\}$.
	\end{enumerate}
\end{theorem}
\textit{Proof:} \\
1. For the noise variance prediction $\hat{\sigma}^{2}$ it holds that:

\begin{align*}\hat{\sigma}^{2}_{\pi}=g_{NV}(g_{D}(\mathcal{D}_{\pi}),g_k(g_{D}(\mathcal{D}_{\pi}),\mathcal{V}_{\mathcal{S}}))=g_{NV}(g_{D}(\mathcal{D}),g_k(g_{D}(\mathcal{D}),\mathcal{V}_{\mathcal{S}}))=\hat{\sigma}^{2}\end{align*}
as $g_{D}(\mathcal{D})=g_{D}(\mathcal{D}_{\pi})$ according to Theorem \ref{dataset_enc_theorem}. For the kernel parameter prediction $\hat{\theta}_{\mathcal{S}}$ it holds similarly
\begin{align*}
\hat{\theta}_{\mathcal{S}}^{\pi}=g_{\Theta_{\mathcal{S}}}(g_k(g_{D}(\mathcal{D}_{\pi}),\mathcal{V}_{\mathcal{S}}),\mathcal{B}_{\mathcal{S}})=g_{\Theta_{\mathcal{S}}}(g_k(g_{D}(\mathcal{D}),\mathcal{V}_{\mathcal{S}}),\mathcal{B}_{\mathcal{S}})=\hat{\theta}_{\mathcal{S}}
\end{align*}
as $g_{D}(\mathcal{D})=g_{D}(\mathcal{D}_{\pi})$ according to Theorem \ref{dataset_enc_theorem}.

2. Let $\hat{\pi}$ be a permutation applied to the input dimensions. Let $[\hat{\mathcal{V}}_{1},\dots,\hat{\mathcal{V}}_{d}]$ be the output of the kernel encoder-decoder for the permuted input and $[\tilde{\mathcal{V}}_{1},\dots,\tilde{\mathcal{V}}_{d}]$ for the unpermuted input.  As $\hat{B}_{j}^{(i)}=B_{j}^{(\hat{\pi}(i))}$ (by definition) and $\hat{v}_{j}^{(i)}=\tilde{v}_{j}^{(\hat{\pi}(i))}$ by Theorem \ref{kernel_enc_dec_theorem}  with $\hat{v}_{j}^{(i)}\in \hat{\mathcal{V}}_{i}$ and $\tilde{v}_{j}^{(i)} \in \tilde{\mathcal{V}}_{i}$  it follows
\begin{align*}
\hat{\theta}_{\hat{B}_{j}^{(i)}}=\textbf{MLP}_{\hat{B}_{j}^{(i)}}(\hat{v}_{j}^{(i)}) = \textbf{MLP}_{B_{j}^{(\hat{\pi}(i))}}(\tilde{v}_{j}^{(\hat{\pi}(i))}) =\theta_{B_{j}^{(\hat{\pi}(i))}}.
\end{align*}

3. Let $i\in\{1,\dots,d\}$ and let $\tilde{\pi}$ be a permutation of the base symbols in dimension $i$. Let $\tilde{\mathcal{V}}_{i}=[\tilde{v}_{1}^{(i)},\dots,\tilde{v}_{N_{i}}^{(i)}]$ be the output of the kernel-encoder-decoder for the unpermuted input and $\hat{\mathcal{V}}_{i}=[\hat{v}_{1}^{(i)},\dots,\hat{v}_{N_{i}}^{(i)}]$ for the permuted input. As $\hat{B}_{j}^{(i)}=B_{\tilde{\pi}(j)}^{(i)}$ (by definition) and $\hat{v}_{j}^{(i)}=\tilde{v}_{\tilde{\pi}(j)}^{(i)}$ by Theorem \ref{kernel_enc_dec_theorem} it follows
\begin{align*}
\hat{\theta}_{\hat{B}_{j}^{(i)}}=\textbf{MLP}_{\hat{B}_{j}^{(i)}}(\hat{v}_{j}^{(i)}) = \textbf{MLP}_{B_{\tilde{\pi}(j)}^{(i)}}(\tilde{v}_{\tilde{\pi}(j)}^{(i)}) =\theta_{B_{\tilde{\pi}(j)}^{(i)}}.
\end{align*}

4.  Let $\hat{\pi}$ be a permutation applied to the input dimensions. Let $[\mathbf{h}_{i}]_{i=1}^{d}$ be the output of the dataset-encoder for the unpermuted input dataset $\mathcal{D}$ and let $[\hat{\mathbf{h}}_{i}]_{i=1}^{d}$ be the output of the dataset-encoder for the permuted input dataset $\mathcal{D}_{\hat{\pi}}$. Furthermore, let $\tilde{\mathcal{V}}_{\mathcal{S}_{\hat{\pi}}}$ be the output of the kernel-encoder-decoder for the permuted expression $\mathcal{S}_{\hat{\pi}}$. Then, it holds for the noise variance prediction:
\begin{align*}
\hat{\sigma}^{2}_{\hat{\pi}}=g_{NV}([\hat{\mathbf{h}}_{i}]_{i=1}^{d},\tilde{V}_{\mathcal{S}_{\hat{\pi}}})&=\textbf{MLP}_{W}\bigg(\textbf{Concat}\bigg(\textbf{MeanAgg}([\hat{\mathbf{h}}_{i}]_{i=1}^{d}),\textbf{MeanAgg}(\tilde{V}_{\mathcal{S}_{\hat{\pi}}})\bigg)\bigg)\\
&=\textbf{MLP}_{W}\bigg(\textbf{Concat}\bigg(\textbf{MeanAgg}([\mathbf{h}_{i}]_{i=1}^{d}),\textbf{MeanAgg}(\tilde{V}_{\mathcal{S}})\bigg)\bigg)\\
&=g_{NV}([\mathbf{h}_{i}]_{i=1}^{d},\tilde{V}_{\mathcal{S}})=\hat{\sigma}^{2},
\end{align*}
since $\textbf{MeanAgg}([\hat{\mathbf{h}}_{i}]_{i=1}^{d})=\textbf{MeanAgg}([\mathbf{h}_{i}]_{i=1}^{d})$ and $\textbf{MeanAgg}(\tilde{V}_{\mathcal{S}_{\hat{\pi}}})=\textbf{MeanAgg}(\tilde{V}_{\mathcal{S}})$ as the mean aggregation is permutation invariant. For the permutation $\tilde{\pi}$ of the base symbols in dimension $i$ the proof goes accordingly via permuted output $\tilde{\mathcal{V}}_{\mathcal{S}_{\tilde{\pi}}}$ of the kernel-encoder-decoder
and
\begin{align*}
\hat{\sigma}^{2}_{\tilde{\pi}}=g_{NV}([\mathbf{h}_{i}]_{i=1}^{d},\tilde{V}_{\mathcal{S}_{\tilde{\pi}}})&=\textbf{MLP}_{W}\bigg(\textbf{Concat}\bigg(\textbf{MeanAgg}([\mathbf{h}_{i}]_{i=1}^{d}),\textbf{MeanAgg}(\tilde{V}_{\mathcal{S}_{\hat{\pi}}})\bigg)\bigg)\\
&=\textbf{MLP}_{W}\bigg(\textbf{Concat}\bigg(\textbf{MeanAgg}([\mathbf{h}_{i}]_{i=1}^{d}),\textbf{MeanAgg}(\tilde{V}_{\mathcal{S}})\bigg)\bigg)\\
&=g_{NV}([\mathbf{h}_{i}]_{i=1}^{d},\tilde{V}_{\mathcal{S}})=\hat{\sigma}^{2}.
\end{align*}
$\blacksquare$
\begin{figure}[t]
	\centering
	\includegraphics[width=0.7\linewidth]{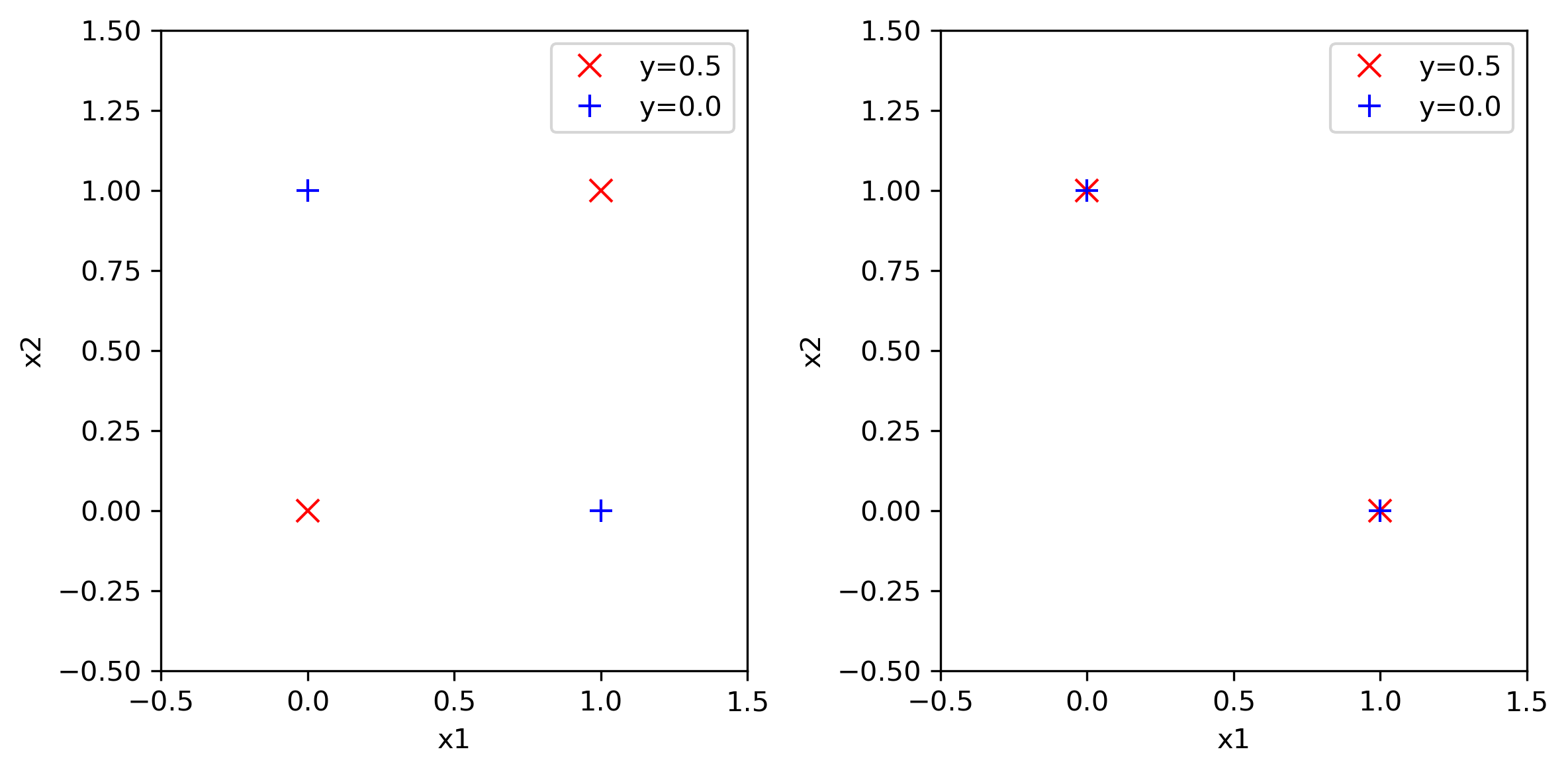}
	\caption[Negative impact of dimension specific invariance.]{In this figure, we see two datasets with four datapoints each. Both datasets would be treated equivalently by the dataset encoder if it would be invariant to a permutation $\pi$ of a single dimension-wise input sequence $\mathcal{H}^{(i)}=[(x_{j}^{(i)},y_{j})]_{j=1}^{n}$ for some $i\in\{1,\dots,d\}$. For the second dataset, we swap the values of the first dimension of the red datapoints, thus doing a permutation of a single dimension-wise input sequence. Considering the $y$ values, the second dataset has high noise, while the first dataset could also come from a low noise process. An invariant network would output the exact same GP parameters for both datasets, a clearly undesirable property.}
	\label{fig:invarianceplot}
\end{figure}

\subsection{Reason for Change of Dataset Encoder}
In this section, we analyze a property that motivates the change to the architecture of the dataset encoder compared to \cite{AHGP}. We consider a permutation $\pi$ of a single dimension-wise input sequence $\mathcal{H}^{(i)}=[(x_{j}^{(i)},y_{j})]_{j=1}^{n}$ for some $i\in\{1,\dots,d\}$. Our architecture is in general not invariant to such a permutation, in contrast to the dataset encoder in \cite{AHGP} which would output the same kernel parameters also after a permutation of a single dimension-wise input sequence. In Figure \ref{fig:invarianceplot} we show two datasets that would be treated exactly as the same dataset in case such an invariance exists - the first dataset might come from a low noise process while the second dataset clearly contains large noise. Thus, assigning the same GP parameters to both datasets is undesirable in this case. Our trained amortization network, equipped with an $\mathrm{SE}$ kernel in each dimension, predicts for the left dataset a noise value of $\hat{\sigma}_{\mathcal{D}_{1}}=0.18$ and for the right dataset $\hat{\sigma}_{\mathcal{D}_{2}}=0.213$. This shows that our network is not invariant to such a permutation and also assigns higher noise to the second dataset.
\end{document}